\documentclass[10pt, twoside, openright]{scrartcl}

\usepackage[ngerman, english]{babel}
\usepackage[a4paper,%
            top=2.5cm,bottom=3.5cm,right=2.5cm,left=2.5cm,bindingoffset=5mm]{geometry}

\usepackage[utf8]{inputenc}
\usepackage[T1]{fontenc}

\usepackage{amsmath}    % math packages
\usepackage{amssymb}

\usepackage{dsfont}     % for one-vector etc.
\usepackage{amsfonts}

\usepackage{graphicx}   % package to include graphics
\usepackage{subfigure}  % package needed for sub-captions a) b) etc.

\usepackage[sc]{mathpazo}   % palatino font
\usepackage{booktabs}   % used for nice tables
\usepackage{dcolumn}    % used for tables centered around decimal point
\usepackage{bm}         % bold Greek letters. Usage $\bm{\beta}$

\usepackage{colortbl}   % needed for color in tables

\usepackage[authoryear,longnamesfirst]{natbib}     % bibliography
\bibpunct{(}{)}{;}{a}{}{,}

\usepackage{setspace}
\usepackage{rotating}

\usepackage{ulem}      % use this for sout
\usepackage{fancyhdr}
\usepackage{fancyvrb}  % fancy verbatim

\usepackage{%
  ellipsis, % Korrigiert den Weißraum um Auslassungspunkte
  ragged2e, % Ermöglicht Flattersatz mit Silbentrennung
 marginnote,% Für bessere Randnotizen mit \marginnote statt
}
\usepackage{keyval}

\usepackage{wallpaper}   % needed for titlepage

\usepackage{float}

\usepackage{framed}
\definecolor{shadecolor}{cmyk}{0.02,0.02,0.02,0.02}

\usepackage[toc,page]{appendix}

\RequirePackage{fancyvrb}
\DefineVerbatimEnvironment{Sinput}{Verbatim}{fontsize=\footnotesize}
\DefineVerbatimEnvironment{Soutput}{Verbatim}{fontsize=\footnotesize}

\setcapindent{2em}
\addtokomafont{caption}{\small}
\setkomafont{captionlabel}{\bfseries}

\newcommand{\COMMENT}[1]{}

\newcommand{\thres}{\pi_{\text{thr}}}

\DeclareMathOperator{\PCER}{\mathit{PCER}}
\DeclareMathOperator{\PFER}{\mathit{PFER}}
\DeclareMathOperator{\PFERmax}{\mathit{PFER}_{\text{max}}}
\DeclareMathOperator{\FWER}{\mathit{FWER}}
\DeclareMathOperator{\FDR}{\mathit{FDR}}

\DeclareMathOperator{\Sstable}{\hat{S}_{\text{stable}}}

\makeatletter
\newcommand\floatc@simplerule[2]{{\@fs@cfont #1} #2\par}
\newcommand\fs@simplerule{\def\@fs@cfont{\bfseries}\let\@fs@capt\floatc@simplerule
  \def\@fs@pre{\hrule height1.2pt depth0pt \kern4pt}%
  \def\@fs@mid{\vspace*{0.5em} \hrule height.3pt depth0pt \vspace*{0.8em} \kern4pt}%
  \def\@fs@post{\kern4pt \hrule height1.2pt depth0pt \kern4pt \relax}%
  \let\@fs@iftopcapt\iftrue}
\makeatother

\floatstyle{simplerule}
\newfloat{algorithm}{thp}{alg}
\floatname{algorithm}{Algorithm}

\renewcommand{\u}{\mathbf{u}}

\newcommand{\x}{\mathbf{x}}

\newcommand{\X}{\mathbf{X}}

\newcommand{\I}{\mathbf{I}}
\newcommand{\y}{\mathbf{y}}

\makeatletter
\def\sprachtestE{Abstract}

\if@titlepage
  \renewenvironment{abstract}{%
      \titlepage
      \null\vfil
      \@beginparpenalty\@lowpenalty
      \begin{flushleft}%
      	\vspace*{-55\p@}
        \LARGE \bfseries \abstractname
        \@endparpenalty\@M
      \end{flushleft}}%
      {\ifx\abstractname\sprachtestE
      \par\vfil\@Keywords
      \else
      \par\vfil\@Schlagwoerter
      \fi
      \endtitlepage}
\else
  \renewenvironment{abstract}{%
      \if@twocolumn
        \chapter*{\abstractname}%
      \else
        \small
        \begin{flushleft}%
          {\LARGE\bfseries \abstractname\vspace{-.5em}\vspace{\z@}}%
        \end{flushleft}%
        \quotation
      \fi}
      {\if@twocolumn\else\endquotation\fi}
\fi
\makeatother

\makeatletter
\newcommand*{\Anrede}[1]{\gdef\@Anrede{#1}}
\newcommand*{\Nachname}[1]{\gdef\@Nachname{#1}}
\newcommand*{\Dateiname}[1]{\gdef\@Dateiname{#1}}
\newcommand*{\NachnameohneUmlaute}[1]{\gdef\@NachnameohneUmlaute{#1}}
\newcommand*{\Vorname}[1]{\gdef\@Vorname{#1}}
\newcommand*{\Geburtsdatum}[1]{\gdef\@Geburtsdatum{#1}}
\newcommand*{\Geburtsort}[1]{\gdef\@Geburtsort{#1}}

\newcommand*{\Typ}[1]{\gdef\@Typ{#1}}
\newcommand*{\Titel}[1]{\gdef\@Titel{#1}}
\newcommand*{\TitelohneUmlaute}[1]{\gdef\@TitelohneUmlaute{#1}}
\newcommand*{\Untertitel}[1]{\gdef\@Untertitel{#1}}

\newcommand*{\GutachterA}[1]{\gdef\@GutachterA{#1}}
\newcommand*{\GutachterB}[1]{\gdef\@GutachterB{#1}}
\newcommand*{\GutachterC}[1]{\gdef\@GutachterC{#1}}

\newcommand*{\Grad}[1]{\gdef\@Grad{#1}}
\newcommand*{\Fach}[1]{\gdef\@Fach{#1}}
\newcommand*{\Fakultaet}[1]{\gdef\@Fakultaet{#1}}
\newcommand*{\Universitaet}[1]{\gdef\@Universitaet{#1}}
\newcommand*{\Dekan}[1]{\gdef\@Dekan{#1}}
\newcommand*{\Rektor}[1]{\gdef\@Rektor{#1}}

\newcommand*{\Abgabedatum}[1]{\gdef\@Abgabedatum{#1}}
\newcommand*{\Pruefungsdatum}[1]{\gdef\@Pruefungsdatum{#1}}

\newcommand*{\Schlagwoerter}[1]{\gdef\@Schlagwoerter{#1}}
\newcommand*{\Keywords}[1]{\gdef\@Keywords{#1}}

\include{metadata}
\makeatother

\RequirePackage{ifpdf}

\makeatletter
\ifpdf
\RequirePackage[%
	pdftitle={Constrained Regression},
	pdfauthor={Benjamin Hofner},
	pdfpagemode=UseOutlines,
        colorlinks=true,	  % bitte nicht ändern!
	linkcolor=black,	  % bitte nicht ändern!
	filecolor=black,	  % bitte nicht ändern!
	urlcolor=black,		  % bitte nicht ändern!
	citecolor=black,	  % bitte nicht ändern!
	pdftex=true,              % bitte nicht ändern!
	plainpages=false,         % bitte nicht ändern!
	hypertexnames=false,      % bitte nicht ändern!
	pdfpagelabels=true,       % bitte nicht ändern!
	hyperindex=true]{hyperref}% bitte nicht ändern!
\else
\fi
\makeatother

\usepackage[all]{hypcap} % fix links to figures: points to the top of the figure

\usepackage{enumerate}
\usepackage{xcolor}

\begin{document}

\title{Controlling false discoveries in high-dimensional situations}
\subtitle{Boosting with stability selection}

\author{Benjamin Hofner\thanks{E-mail:
    {benjamin.hofner@fau.de}} \footnote{Department of Medical
    Informatics, Biometry and Epidemiology, Friedrich-Alexander-Universit\"at
    Erlangen-N\"urnberg, Waldstra{\ss}e 6, 91054 Erlangen, Germany}
  \addtocounter{footnote}{-1} \and Luigi Boccuto \footnote{Greenwood Genetic
    Center, 113 Gregor Mendel Circle, Greenwood, SC 29646, USA } \and Markus
  Göker\footnote{Leibniz Institute DSMZ -- German Collection of Microorganisms
    and Cell Cultures, Inhoffenstraße 7b, 38124 Braunschweig, Germany} }

\date{\today}

\maketitle

\begin{abstract}
  Modern biotechnologies often result in high-dimensional data sets with much
  more variables than observations ($n \ll p$). These data sets pose new
  challenges to statistical analysis: Variable selection becomes one of the most
  important tasks in this setting. We assess the recently proposed flexible
  framework for variable selection called stability selection. By the use of
  resampling procedures, stability selection adds a finite sample error control
  to high-dimensional variable selection procedures such as Lasso or boosting.
  We consider the combination of boosting and stability selection and present
  results from a detailed simulation study that provides insights into the
  usefulness of this combination. Limitations are discussed and guidance on the
  specification and tuning of stability selection is given. The interpretation
  of the used error bounds is elaborated and insights for practical data
  analysis are given. The results will be used to detect differentially
  expressed phenotype measurements in patients with autism spectrum disorders.
  All methods are implemented in the freely available \textsf{R} package
  \textbf{stabs}.

\end{abstract}

\section{Introduction}

Variable selection is a notorious problem in many applications. The researcher
collects many variables on each study subject and then wants to identify the
variables that have an influence on the outcome variable. This problem becomes
especially pronounced with modern high-throughput experiments where the number
of variables $p$ is often much larger than the number of observations $n$
\citep[e.g., genomics, transcriptomics, proteomics, metabolomics, metabonomics
and phenomics;
see][]{Chaturvedi:genomics:2014,Wang:transcriptomics:2009,Mallick:proteomics:2010,Ludwig:metabolab:2011,Lindon:metabonomics:2003,Groth:phenomics:2008}.
One of the major aims in the analysis of these high-dimensional data sets is to
detect the signal variables $S$, while controlling the number of selected noise
variables $N$. Stepwise regression models are a standard approach to variable
selection in settings with relatively few variables. However, even in this case
this approach is known to be very unstable \citep[see
e.g.,][]{flack1987frequency,austin2004automated,austin2008bootstrap}. Recent
approaches that try to overcome this problem and can also be used in
high-dimensional settings with $n \ll p$ include penalized regression approaches
such as the lasso \citep{Tibshirani:Lasso:1996,Efron:LARS:2004}, elastic net
\citep{Zou:elasticnet:2005}, and boosting
\citep{Frie:Hast:Tibs:stat_view_boosting:2000}, or tree based approaches such as
random forests \citep{Breiman2001,Strobl:2006:BMC-Bioinformatics}. More
recently, \citet{Meinshausen:2010} proposed stability selection, an approach
based on resampling of the data set which can be combined with many selection
procedures and is especially useful in high-dimensional settings. Stability
selection has since been widely used, e.g. for gene regulatory network analysis
\citep{Haury:2012:Stabsel_Generegulation,marbach:2012:Stabsel_Generegulation_NATURE},
in genome-wide association studies \citep{He:2011:GWAS_Stabsel}, graphical
models \citep{Fellinghauer:stabsel:2013,Buhlmann:StabselReview:2014} or even in
ecology \citep{Hothorn:2011:EcoMonographs}. In most publications, stability
selection is used in combination with lasso or similar penalization approaches.
Here, we discuss the combination of stability selection with component-wise
functional gradient descent boosting \citep{buehlmann03} which allows one to
specify competing effects, which are subject to selection, more flexibly. For
details on functional gradient descent boosting, see \citet{buehl:hoth:2007} and
\citet{Hofner:mboost:2014}.

We will provide a short, rather non-technical introduction to boosting in
Section~\ref{sec:boosting}. Stability selection, which controls the per-family
error rate, will be introduced in Section~\ref{sec:stability-selection}, where
we also give an overview on common error rates and some guidance on the choice
of the parameters in stability selection. Section~\ref{sec:empirical-evaluation}
presents an empirical evaluation of boosting with stability selection. In our
case study (Section~\ref{sec:data-analysis}) we will examine autism spectrum
disorder (ASD) patients and compare them to healthy controls using the boosting
approach in conjunction with stability selection. The aim is to detect
differentially expressed phenotype measurements. More specifically, we try to
assess which amino acid pathways differ between healthy subjects and ASD
patients.

\section{A Short Introduction to Boosting}
\label{sec:boosting}

Consider a generalized linear model
\begin{equation}\label{eq:lm}
  \mathds{E}(y|\x) = h(\eta(\x))
\end{equation}
with outcome $y$, appropriate response function $h$ and linear predictor
$\eta(x)$. Let the latter be defined as
\begin{equation}
  \label{eq:4}
  \eta(\x) = \beta_0 + \sum_{j = 1}^p \beta_j x_{j},
\end{equation}
with covariates $\x = (x_1, \ldots, x_p)$, and corresponding effects $\beta_j,\
j = 0, \ldots, p$. Model fitting aims at minimizing the expected loss
$\mathds{E}(\rho(y, \x))$ with an appropriate loss function $\rho(y, \x)$. The
loss function is defined by the fitting problem at hand. Thus, for example,
Gaussian regression models, i.e.\ least squares regression models, aim to
minimize the squared loss $\rho(y, \x) = (y - \eta(\x))^2$. Generalized linear
models can be obtained by maximizing the log-likelihood or, analogously, by
minimizing the negative log-likelihood function. Logistic regression models with
binary outcome, for example, can be obtained by using the negative binomial
log-likelihood
\begin{equation*}
  \rho(y, \x) = -y \log(P(y = 1 | \x)) +
  (1 - y) \log (1 - P(y = 1 | \x))
\end{equation*}
as loss function or a reparametrization thereof \citep{buehl:hoth:2007}.

In practice, one cannot minimize the expected loss function. Instead, we
optimize the empirical risk function
\begin{equation}
  \label{eq:2}
  \mathcal{R}(\y, \X) = n^{-1} \sum_{i=1}^n \rho(y_i, \eta(\x_i))
\end{equation}
with observations $\y = (y_1, \ldots, y_n)^\top$ and $\X = (\x^\top_1, \ldots,
\x^\top_n)^\top$. This can be done for arbitrary loss functions by
component-wise functional gradient descent boosting \citep{buehlmann03}. The
algorithm is especially attractive owing to its intrinsic variable selection
properties \citep{Kneib:Hothorn:Tutz:modelchoice:2009,Hofner:unbiased:2011}.

One begins with a constant model $\hat{\eta}^{[0]}(\x_i) \equiv 0$ and computes the
residuals $\u^{[1]} = (u_1^{[1]} , \ldots, u_n^{[1]})^\top$ defined by the
negative gradient of the loss function
\begin{equation}\label{eq:1}
  u_i^{[m]} := - \left. \frac{\partial \rho(y_i, \eta)}{\partial \eta} \right|_{\eta =
    \hat{\eta}^{[m-1]}(\x_i)}
\end{equation}
evaluated at the fit of the previous iteration $\hat{\eta}^{[m-1]}(\x_i)$
\citep[see][]{buehlmann03,buehl:hoth:2007,Hothorn+Buehlmann+Kneib+Schmid+Hofner:mboost:2010}.
Each variable $x_1 , \ldots, x_p$ is fitted separately to the residuals
$\u^{[m]}$ by least squares estimation (this is called the ``base-learner''),
and only the variable $j^*$ that describes these residuals best is updated by
adding a small percentage $\nu$ of the fit $\hat{\beta}_{j^*}$ (e.g., $\nu =
10\%$) to the current model fit, i.e.,
\begin{equation*}
  \hat{\eta}^{[m]} = \hat{\eta}^{[m-1]} + \nu \cdot \hat{\beta}_{j^*}.
\end{equation*}
New residuals $\u^{[m+1]}$ are computed, and the whole procedure is iterated
until a fixed number of iterations $m = m_{\text{stop}}$ is reached. The final
model $\hat{\eta}^{[m_{\text{stop}}]}(\x_i)$ is defined as the sum of all models
fitted in this process. Instead of using linear base-learners (i.e., linear
effects) to fit the negative gradient vector $\u^{[m]}$ in each boosting step,
one can also specify smooth base-learners for the variables $x_j$ \cite[see
e.g.][]{Schmid:Hothorn:boosting-p-Splines}, which are then fitted by penalized
least squares estimation. As we update only one modeling component in each
boosting iteration, variables are selected by stopping the boosting procedure
after an appropriate number of iterations (``early stopping''). This number is
usually determined using cross-validation techniques \cite[see
e.g.,][]{Mayr:mstop:2012}.

\section{Stability Selection}
\label{sec:stability-selection}

A problem of many statistical learning approaches including boosting with early
stopping is that despite regularization one often ends up with relatively rich
models \citep{Mayr:mstop:2012,Meinshausen:2010}. A lot of noise variables might
be erroneously selected. To improve the selection process and to obtain an error
control for the number of falsely selected noise variables
\citet{Meinshausen:2010} proposed stability selection. This is a versatile
approach, which can be combined with all high-dimensional variable selection
approaches. Stability selection is based on sub-sampling and controls the
\emph{per-family error rate} $\mathds{E}(V)$, where $V$ is the number of false
positive variables (for more details on error rates see
Section~\ref{sec:error_rates}).

Consider a data set with $p$ predictor variables $x_j,\, j = 1, \ldots, p$ and
an outcome variable $y$. Let $S \subseteq \{1, \ldots, p\}$ be the set of signal
variables, and let $N \subseteq \{1, \ldots, p\} / S$ be the set of noise
variables. The set of variables that are selected by the statistical learning
procedure is denoted by $\hat{S}_n \subseteq \{1, \ldots, p\}$. This set
$\hat{S}_n$ can be considered to be an estimator of $S$, based on a data set
with $n$ observations. In short, for stability selection with boosting one
proceeds as follows:

\begin{enumerate}[\quad1.)]\itemsep0pt
\item \label{item:subset} Select a random subset of size $\lfloor n/2 \rfloor$
  of the data, where $\lfloor x \rfloor$ denotes the largest integer $\leq x$.
\item \label{item:modelfit} Fit a boosting model and continue to increase the
  number of boosting iterations $m_{\text{stop}}$ until $q$ base-learners are
  selected. $\hat{S}_{\lfloor n/2 \rfloor,\, b}$ denotes the set of selected variables.
\item Repeat the steps \ref{item:subset}) and \ref{item:modelfit}) for $b = 1,
  \ldots, B$.
\item Compute the relative selection frequencies
  \begin{equation}\label{eq:selprob}
    \hat{\pi}_j := \frac{1}{B} \sum_{b = 1}^B \mathds{I}_{\{j \in \hat{S}_{\lfloor n/2 \rfloor,\, b}\}}
  \end{equation}
  per variable (or actually per base-learner).
\item Select all base-learners that were selected with a frequency of at least
  $\thres$, where $\thres$ is a pre-specified threshold value. Thus, we obtain a
  set of \emph{stable variables} $\Sstable := \{j: \hat{\pi}_j \geq \thres\}$.
\end{enumerate}

\citet{Meinshausen:2010} show that this selection procedure controls the
per-family error rate ($\PFER$). An upper bound is given by
\begin{equation}\label{eq:MB}
  \mathds{E}(V) \leq \frac{q^2}{(2\thres - 1) p}
\end{equation}
where $q$ is the number of selected variables per boosting run, $p$ is the
number of (possible) predictors and $\thres$ is the threshold for selection
probability. The theory requires two assumptions to ensure that the error bound
holds:
\begin{enumerate}[(i)]
\item The distribution $\{\mathds{I}_{\{j \in \Sstable\}}, j \in N\}$ needs to
  be exchangeable for all noise variables $N$.\label{item:exchangeability}
\item The original selection procedure, boosting in our case, must not be worse
  than random guessing.\label{item:random_guessing}
\end{enumerate}
In practice, assumption~\eqref{item:exchangeability} essentially means that each
noise variable has the same selection probability. Thus, all \emph{noise
  variables} should, for example, have the same correlation with the signal
variables (and the outcome). For examples of situations where exchangeability is
given see \citet{Meinshausen:2010}. Assumption~\eqref{item:random_guessing}
means that signal variables should be selected with higher probability than
noise variables. This assumption is usually not very restrictive as we would
expect it to hold for any sensible selection procedure.

\paragraph{Choice of parameters}

The stability selection procedure mainly depends on two parameters: the number
of selected variables per boosting model $q$ and the threshold value for stable
variables $\thres$. \citet{Meinshausen:2010} propose to chose $\thres \in (0.6,
0.9)$ and claim that the threshold has little influence on the selection
procedure. In general, any value $\in (0.5, 1)$ is potentially acceptable, i.e.\
a variable should be selected in more than half of the fitted models in order to
be considered stable. The number of selected variables $q$ should be chosen so
high that in theory all signal variables $S$ can be chosen. If $q$ was too
small, one would inevitably select only a small subset of the signal variables
$S$ in the set $\Sstable$ as $|\Sstable| \leq |\hat{S}_{\lfloor n/2 \rfloor,\,
  b}| = q$ (if $\thres > 0.5$).

The choice of the number of subsamples $B$ is of minor importance as long as it
is large enough. \citet{Meinshausen:2010} propose to use $B = 100$ replicates,
which seems to be sufficient for an accurate estimation of $\hat{\pi}_j$ in most
situations.

In general, we would recommend to choose an upper bound $\PFERmax$ for the
$\PFER$ and specify either $q$ or $\thres$, preferably $\thres$. The missing
parameter can then be computed from Equation~(\ref{eq:MB}), where equality is
assumed. In a second step, one should check that the computed value is sensible,
i.e.\ that $\thres \in (0.5, 1)$, or that $q$ is not too small, or that
$\PFERmax$ is not too small or too large. Note that the $\PFER$ can be greater
than one as it resembles the tolerable expected number of falsely selected noise
variables. An overview on common error rates is given in
Section~\ref{sec:error_rates}, where we also give some guidance on the choice of
$\PFERmax$.

The size of the subsamples is no tuning parameter but should always be chosen to
be $\lfloor n/2 \rfloor$. This an essential requirement for the derivation of
the error bound~(\ref{eq:MB}) as can be seen in the proof of Lemma 2
\citep{Meinshausen:2010}, which is used to proof the error bound. Other (larger)
subsample sizes would theoretically be possible but would require the derivation
of a different error bound for that situation.

\subsection{Improved Version of Stability Selection}

A modification of stability selection was introduced by
\citet{ShahSamworth:Stabsel:2013}. One major difference to the original
stability selection approach is that instead of using $B$ independent subsamples
of the data, \citeauthor{ShahSamworth:Stabsel:2013} use $2B$ complementary
pairs: One draws $B$ subsamples of size $\lfloor n/2 \rfloor$ from the data and
uses, for each subsample, the remaining observations as a second complementary
subsample.

More importantly, error bounds are theoretically derived that hold without
assuming exchangeability of the noise variables (and without assuming that the
original selection procedure is not worse than random guessing). The drawback of
being able to drop these assumptions~\eqref{item:exchangeability}
and~\eqref{item:random_guessing}) is that the modified bounds do not control the
per-family error rate, but the \emph{expected number of selected variables with
  low selection probability}
\begin{equation}\label{eq:exp_low_sel}
  \mathds{E}(|\Sstable \cap L_\theta|),
\end{equation}
where $\Sstable$ denotes the set of variables selected by stability selection,
and $L_\theta = \{j: \hat{\pi}_j \leq \theta \}$ denotes the set of variables
that have a low selection probability under $\hat{S}_{\lfloor n/2 \rfloor}$,
i.e.\ a selection probability below $\theta$ in one boosting run on a subsample
of size $\lfloor n/2 \rfloor$. Usually, this threshold for low selection
probabilities is chosen as $\theta = \frac{q}{p}$, i.e.\ the average fraction of
selected variables. Thus, this error rate represents the expected number of
variables that are unlikely to be selected but are selected.

Here, the selection probability $\hat{\pi}_j$ (Eq.~\ref{eq:selprob}) needs to be
computed over all $2B$ random (complementary) subsamples. Additionally, let the
simultaneous selection probability $\widetilde{\pi}_j$ be defined as follows
\citep{ShahSamworth:Stabsel:2013}:
\begin{equation}\label{eq:simselprob}
  \widetilde{\pi}_j := \frac{1}{B} \sum_{b = 1}^B \mathds{I}_{\{j \in
    \hat{S}^{1}_b \}} \cdot \mathds{I}_{\{j \in
    \hat{S}^{2}_b \}},
\end{equation}
where $\mathds{I}_{\{j \in S\}}$ is the indicator function which is one if $j
\in S$ and zero otherwise. $\hat{S}^{1}_b$ is the set of selected variables on
the $b$th random subset of size $\lfloor n/2 \rfloor$ and $\hat{S}^{2}_b$ is the
selection on the complementary pair of this random subset. Note that both sets
of selected variables are derived with the original learning procedure without
applying the stability selection threshold so far.

\citet{ShahSamworth:Stabsel:2013} derive three error bounds for the
\emph{expected number of low selection probability variables}:
\begin{enumerate}[(E1)]
\item \label{item:errorbound1} A worst case error bound is derived for all $\thres \in
  (0.5, 1]$:
  \begin{equation}\label{eq:errorbound1}
    \mathds{E}(|\Sstable \cap L_\theta|) \leq
    \frac{\theta}{2\thres - 1} \; \mathds{E}(|\hat{S}_{\lfloor n/2 \rfloor} \cap
      L_\theta|) \leq \frac{\theta}{2\thres - 1} \; q
  \end{equation}
  If $\theta = \frac{q}{p}$, this error bound is equal to (\ref{eq:MB}) but does
  not require that assumptions~\eqref{item:exchangeability}
  and~\eqref{item:random_guessing} hold.
\item \label{item:errorbound2} A second, tighter, error bound assumes that the
  simultaneous selection probabilities $\widetilde{\pi}_j$ have a unimodal
  probability distribution for all $j \in L_\theta$. If additionally $\theta
  \leq 1 / \sqrt{3} \approx 0.577$ holds, the error bound can be written as
  \begin{equation}\label{eq:errorbound2}
    \mathds{E}(|\Sstable \cap L_\theta|) \leq
    \frac{\theta}{c(\thres, B)} \; \mathds{E}(|\hat{S}_{\lfloor n/2 \rfloor} \cap
    L_\theta|) \leq \frac{\theta}{c(\thres, B)} \; q
  \end{equation}
  with constant
  \begin{equation*}
    c(\thres, B) =
    \begin{cases}
      2 \left( 2\thres - 1 - \frac{1}{2B} \right) &
        \text{\quad if } \thres \in (c_{\text{min}}, \frac{3}{4}]\\
      \frac{1 + 1/B}{4 \left( 1 - \thres + \frac{1}{2B} \right) } &
        \text{\quad if } \thres \in (\frac{3}{4}, 1],
    \end{cases}
  \end{equation*}
  and $c_{\text{min}} = \min(\frac{1}{2} + \theta^2, \frac{1}{2} + \frac{1}{2B}
  + \frac{3}{4} \theta^2)$. One needs to further assume that $\thres \in
  \Bigl\{\frac{1}{2} + \frac{2}{2B}, \frac{1}{2} + \frac{3}{2B}, \ldots, 1
  \Bigr\}$ for the bound to hold. However, this is no restriction in practice,
  as for typical values of $B$ such as $B = 50$ or $B = 100$, all values of
  $\thres \geq 0.51$ in steps of $0.01$ or $\thres \geq 0.505$ in steps of
  $0.005$, respectively, are permitted.
\item \label{item:errorbound3}The third error bound assumes that the
  simultaneous selection probabilities $\widetilde{\pi}_j$ have an r-concave
  probability distribution with $r = -\frac{1}{2}$ and that the selection
  probabilities $\hat{\pi}_j$ have an r-concave probability distribution with $r
  = -\frac{1}{4}$ for all $j \in L_\theta$. With $f_j$ being the distribution of
  $\widetilde{\pi}_j$ and $g_j$ being the distribution of $\hat{\pi}_j$ this is
  equivalent to the assumptions that $f_j^{-1/2}$ and $g_j^{-1/4}$ must be
  convex. The r-concavity assumption lies in between unimodality and the
  stronger log-concavity assumption. For details on r-concavity we refer to
  \citet{ShahSamworth:Stabsel:2013}. If the r-concavity assumption holds, the
  error bound can be further refined as
  \begin{eqnarray}
      \mathds{E}(|\Sstable \cap L_\theta|) & \leq & \min \left\{
        D\left(2 \thres - 1; \theta^2, B, -\frac{1}{2}\right),
        D\left(\thres; \theta, 2B, -\frac{1}{4}\right)
      \right\} \; |L_\theta| \nonumber\\
      & \leq & \min \left\{
        D\left(2 \thres - 1; \theta^2, B, -\frac{1}{2}\right),
        D\left(\thres; \theta, 2B, -\frac{1}{4}\right)
      \right\} \; p.\label{eq:errorbound3}
  \end{eqnarray}
  The function $D(\xi; \theta, B, r)$ denotes the maximum of the probability
  $P(X \leq \xi)$ with $\mathds{E}(X) \leq \theta$ over all $r$-concave random
  variables $X$ on a discrete support $\{0, 1/B, 2/B, \ldots, 1\}$. For details
  see \citet[Appendix A.4]{ShahSamworth:Stabsel:2013}.
\end{enumerate}
With these additional assumptions we get much tighter error bounds. The reason
for tighter bounds can be found in the application of refined bounds in Markov's
inequality that make use of the distributional assumptions. Markov's inequality
is used on the simultaneous selection probabilities $\widetilde{\pi}_j$ in the
derivation of the error bounds
\citep[see][App.~A.1--A.3]{ShahSamworth:Stabsel:2013}.

One should be aware that the assumptions are on the \emph{distribution} of the
selection probabilities and not on the selection probability itself. The
unimodality assumption seems to generally hold in practice. The r-concavity
assumption may fail, if the number of subsamples $B$ increases, since as $B$
increases, r-concavity requires an increasing number of inequalities to hold for
the distribution of $\widetilde{\pi}_j$. However, the same problem does not
occur for the unimodal bound, and when $B = 50$, the bounds constructed using
the r-concavity assumption seem to hold in a wide variety of scenarios (Shah,
2014, personal communication; see also Section~\ref{sec:results_sim}).

\paragraph{Interpretation of Error Bounds~(E\ref{item:errorbound1}) -- (E\ref{item:errorbound3})}

If the exchangeability assumption holds and the selection procedure is not worse
than random guessing, then all noise variables have a ``below average''
selection probability. Hence, the low selection probability variables will
include all noise variables, i.e.\ $L_\theta = N$. Controlling the
\emph{expected number of selected variables with low selection probability} is
thus in this case identical to controlling the expected number of false
positives:
\begin{equation*}
  \mathds{E}(|\Sstable \cap L_\theta|) = \mathds{E}(|\Sstable \cap N|) = \mathds{E}(V).
\end{equation*}
Stability selection can consequently be thought to control the per-family error
rate in all three cases (E\ref{item:errorbound1}) -- (E\ref{item:errorbound3}).
On the other hand, if exchangeability does not hold, this means that we have
``special'' noise variables, e.g., noise variables that are stronger correlated
with signal variables than other noise variables. If this correlation is so
strong that a variable is selected with ``above average selection probability'',
it is difficult to think of this variable as noise variables anyway. Thus
controlling the \emph{expected number of selected variables with low selection
  probability} is again similar or even practically identical to controlling the
expected number of false positives.

\subsection{Definitions and Discussion of Common Error Rates}\label{sec:error_rates}

There are various definitions of error rates that are used in statistics,
especially in the case of multiple testing. Let $m$ be the number of tested
hypothesis, $R$ the number of rejected hypothesis and $V$ the number falsely
rejected hypotheses as defined above \citep[cf.][]{BenjaminiHochberg:FDR:1995}.
In our case, $m$ is the number of predictor variables $p$ or more general the
number of base-learners in the boosting model. Commonly used error rates include
the per-comparison error rate $\PCER = \mathds{E}(V)/m$, the per-family error
rate $\PFER = \mathds{E}(V)$, the family-wise error rate $\FWER = \mathds{P}(V
\geq 1)$, and the false discovery rate $\FDR = \mathds{E}(\frac{V}{R})$
\citep{BenjaminiHochberg:FDR:1995}. The per-comparison error rate is the
standard error rate without adjustment for multiplicity.

For a given test situation it holds that
\begin{equation*}
  \PCER \leq \FWER \leq \PFER.
\end{equation*}
Thus, for a fixed significance level $\alpha$ it holds that $\PFER$-control is
more conservative than $\FWER$-control which is in turn more conservative than
$\PCER$-control \citep{Dudoit2003}. The $\FDR$, which is often used in (very)
high-dimensional settings such as gene expression studies uses another error
definition by relating the number of false discoveries to the number of rejected
null hypotheses. One can show that in a given test situation
\begin{equation*}
  \FDR \leq \FWER,
\end{equation*}
and thus for a fixed level $\alpha$, $\FWER$-control is more conservative than
$\FDR$-control \citep{Dudoit2003}. In conclusion, it holds that $\FDR \leq \FWER
\leq \PFER$. Controlling the $\PFER$ is a (very) conservative approach for
controlling errors in multiple testing situations. Hence, a procedure that
controls the $\PFER$ at a certain level $\alpha$ also controls all other error
rates discussed in this section at this level. Obviously the error bound will be
very conservative upper bound for both the $\FWER$ and $\FDR$.

The standard approach for hypotheses testing, neglecting multiplicity, would be
to specify a bound for the per-comparison error rate by using a significance
level $\alpha$, e.g. $\alpha = 0.05$. This is equal to specifying $\PFERmax \leq
m \alpha$. This provides some guidance on how to choose an upper bound for the
$\PFER$: Usually, $\alpha \leq \PFERmax \leq m \alpha$ seems a good choice,
where $\PFERmax = \alpha$ would (conservatively) control the $\FWER$ on the
level $\alpha$, while $\PFERmax = m \alpha$ would control the unadjusted
per-comparison error rate on the level $\alpha$. Everything in between can be
considered to control the $\PCER$ on the level $\alpha$ ``with some multiplicity
adjustment''.

\section{Empirical Evaluation}
\label{sec:empirical-evaluation}

To evaluate the impact of the tuning parameters $q$ and $\thres$, the upper
bound $\PFERmax$, and the assumptions for the computation of the upper bound on
the selection properties, we conducted a simulation study using boosting in
conjunction with stability selection. Additionally, we examined the impact of
the characteristics of the data set on the performance.

We considered a classification problem with a binary outcome variable. The data
were generated according to a linear logistic regression model with linear
predictor $\eta = \X\beta$ and
\begin{equation*}
  Y \sim \text{Binom}\left(\frac{\exp(\eta)}{1 + \exp(\eta)}\right).
\end{equation*}
The observations $x_i = (x_{i1}, \ldots, x_{ip}),\; i = 1, \ldots, n$ were
independently drawn from
\begin{equation*}
  x \sim \mathcal{N}(0, \Sigma),
\end{equation*}
and gathered in the design matrix $\X$. We set the number of predictor variables
to $p \in \{100, 500, 1000\}$, and the number of observations to $n \in \{50,
100, 500\}$. The number of influential variables varied within $p_{\text{infl}}
\in \{2, 3, 8\}$, where $\beta_j$ was sampled from $\{-1, 1\}$ for an
influential variable and set to zero for all non-influential variables. We used
two settings for the design matrix:
\begin{enumerate}
\item independent predictor variables, i.e.\ $\Sigma = \I$,
\item and correlated predictor variables drawn from a Toeplitz design with
  covariance matrix $\Sigma_{kl} = 0.9^{|k - l|},\; k,l = 1, \ldots, p$.
\end{enumerate}

For each of the data settings we used all three error bounds in combination with
varying parameters $\thres \in \{0.6, 0.75, 0.9\}$, and $\PFERmax \in \{0.05, 1,
2, 5\}$. We used the standard subsampling scheme with $B = 100$ subsamples for
the error bound~(E\ref{item:errorbound1}) and complementary pairs with $B = 50$
subsamples for the improved error bounds~(E\ref{item:errorbound2}) and
(E\ref{item:errorbound3}). Each simulation setting was repeated 50 times.

\subsection{Results}
\label{sec:results_sim}

\begin{figure}[h]
  \centering
\includegraphics{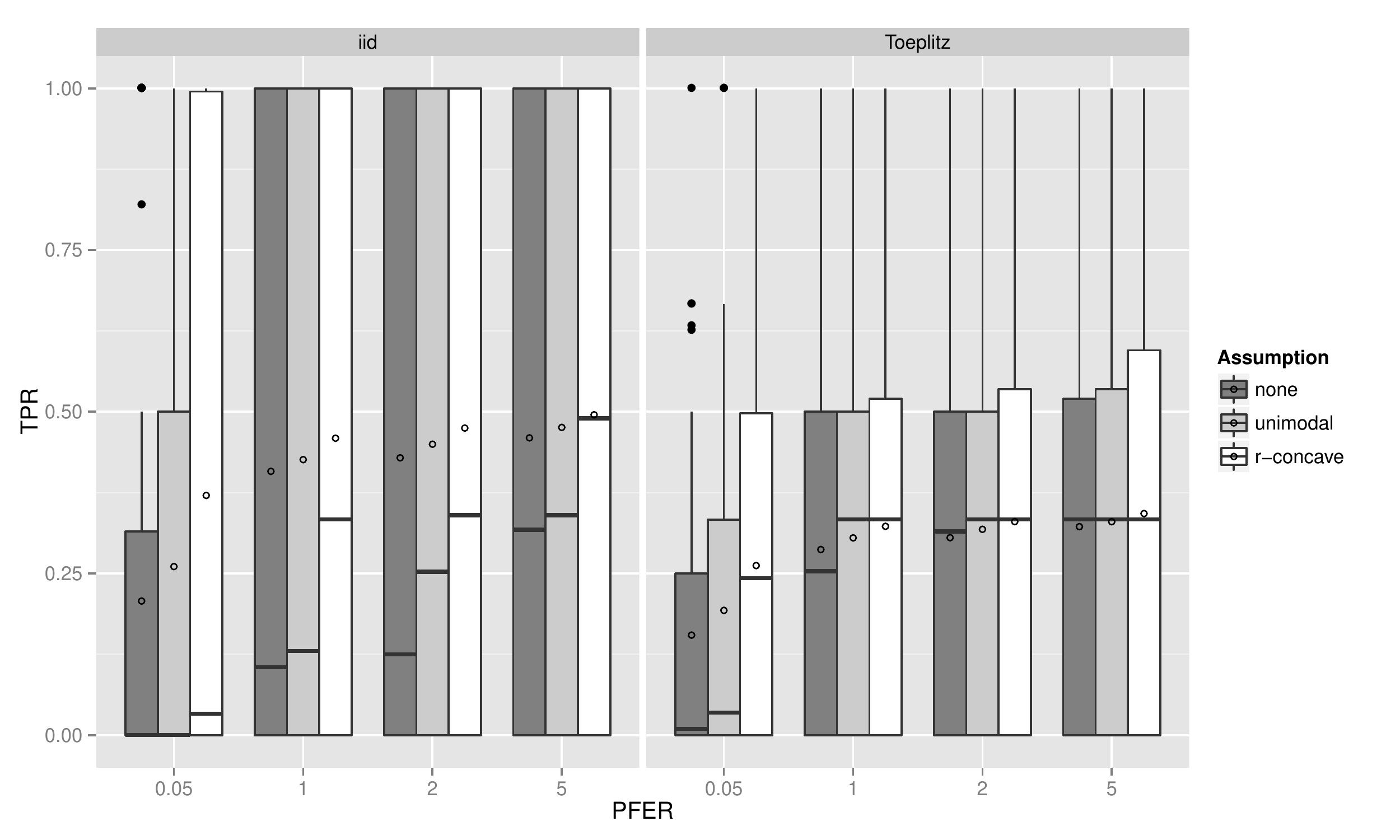}
\caption{Boxplots for the true positives rates (TPR) for all simulation settings
  with separate boxplots for the correlation settings (independent predictor
  variables or Toeplitz design), $\PFERmax$ and the assumption used to compute
  the error bound. Each observation in the boxplot is the average of the 50
  simulation replicates. The open circles in the boxes represent the average
  true positive rates.}
  \label{fig:TPR}
\end{figure}

Figure~\ref{fig:TPR} displays the true positive rates for different $\PFERmax$
bounds, the three assumptions (E\ref{item:errorbound1}) to
(E\ref{item:errorbound3}) and for the two correlation schemes. Different sizes
of the data set ($n$ and $p$) as well as different numbers of true positives
($p_{\text{infl}}$) were not depicted as separate boxplots. For each upper bound
$\PFERmax$ and each data situation (uncorrelated/Toeplitz), the true positive
rate (TPR) increased with stronger assumptions (E\ref{item:errorbound1}) to
(E\ref{item:errorbound3}). The true positive rate was lower when the predictors
were correlated.

\setkeys{Gin}{width = 0.9\textwidth}

%%% Local Variables:
%%% mode: latex
%%% TeX-master: "00_main"
%%% End:

If the number of observations $n$ increased, the TPR increased as well with more
extreme cases for uncorrelated predictors (see Appendix;
Figure~\ref{fig:TPR_by_n}). With very few observations ($n = 50$), the TPR was
generally very small. Considering the size of the subsamples, which is equal to
25, this is quite natural. Recently, \cite{Schmid_etal_PAUC_2012} advocated to
increase the sample size of the subsamples from $\lfloor n/2 \rfloor$ to larger
values to avoid biased selection of base-learners due to too small samples. Yet,
as discussed above, this is currently not possible, as one would need to derive
a different error bound for that situation. Conversely, the TPR decreases with
an increasing number of truly influential variables $p_{\text{infl}}$ (see
Appendix; Figure~\ref{fig:TPR_by_pinfl}). The threshold $\thres$ is less
important (see Appendix; Figure~\ref{fig:TPR_by_cutoff}), as long as it is large
enough to result in enough variables $q$ to be selected and not too large so
that too many variables would be selected in each run (see Appendix;
Figure~\ref{fig:TPR_by_q_and_pinfl}). Note that the dependence on the threshold
is stronger in the case with correlated observations.

\begin{figure}[h]
  \centering
\includegraphics{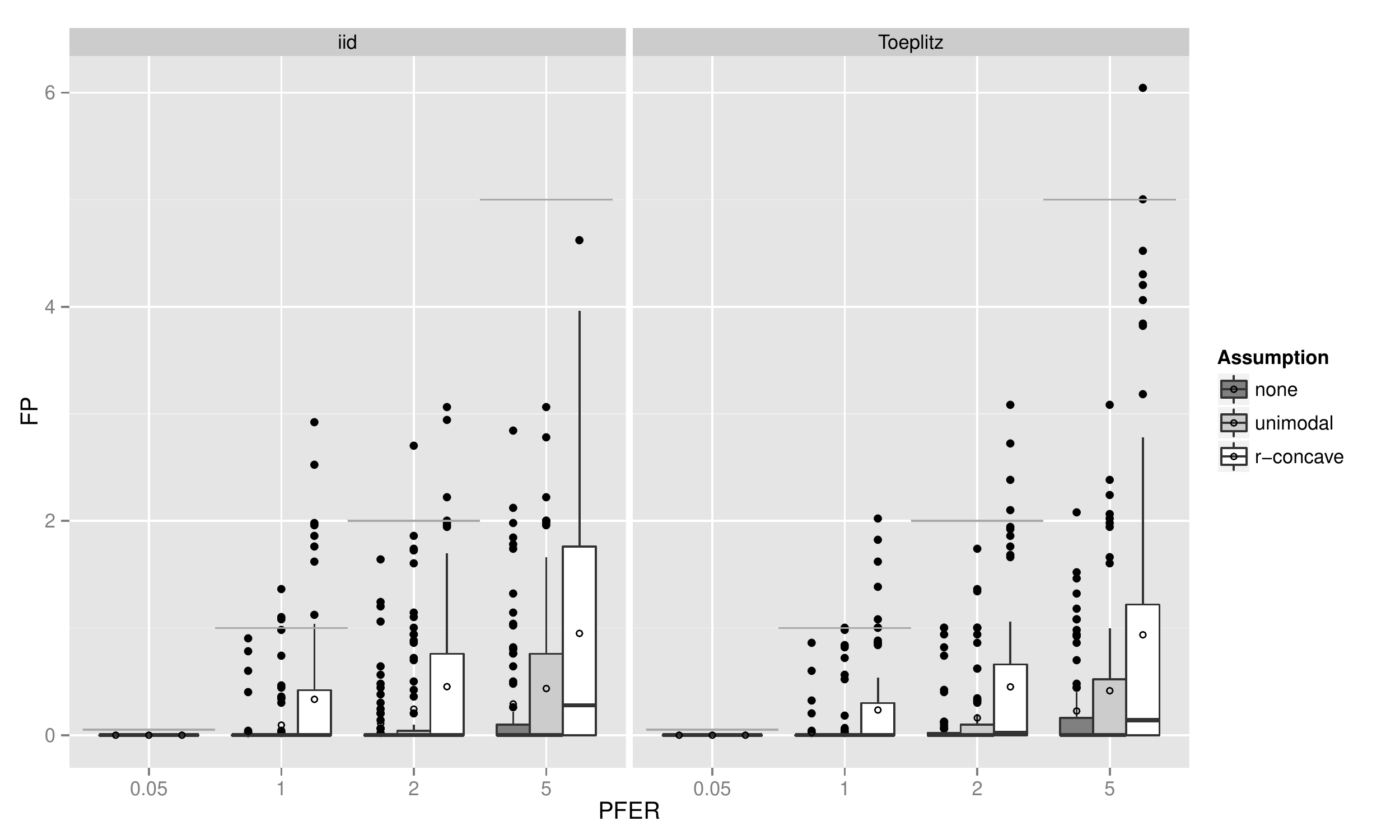}
\caption{Boxplots for the number of false positives for all simulation settings
  with separate boxplots for the correlation settings (independent predictor
  variables or Toeplitz design), $\PFERmax$ and the assumption used to compute
  the error bound. Each observation in the boxplot is the average of the 50
  simulation replicates. The open circles represent the average number of false
  positives. The grey horizontal lines represent the error bounds.}
  \label{fig:FP}
\end{figure}

The number of false positives, which is bounded by the upper bound for the
per-family error rate, is depicted in Figure~\ref{fig:FP}. Overall, the error
rate seemed to be well controlled with some violations of the less conservative
bounds in the median settings. However, overall the error bound was violated in
only 1.2 \% (4 cases) under the unimodality assumption and 4.0 \% (13 cases)
under the r-concavity assumption. Especially the standard error bound
(E\ref{item:errorbound1}) seemed to be conservatively controlled. The average
number of false positives increased with increasing $\PFERmax$ and with stronger
distributional assumptions on the simultaneous selection probabilities. In
general, one should note that stability selection is quite conservative as it
controls the $\PFER$. The given upper bounds for the $\PFER$ corresponded to
per-comparison error rates between $0.05$ and $0.00005$.

\setkeys{Gin}{width = 0.9\textwidth}

%%% Local Variables:
%%% mode: latex
%%% TeX-master: "00_main"
%%% End:

If the number of observations $n$ increased, the number of false positives
decreased on average but the variability increased as well (see Appendix;
Figure~\ref{fig:FP_by_n}). The number of false positives showed a tendency to
decrease with an increasing number of truly influential variables
$p_{\text{infl}}$ (see Appendix; Figure~\ref{fig:FP_by_pinfl}). If the threshold
$\thres$ was larger, i.e., only highly frequently selected variables were
considered to be stable, the number of false positives decreased (see Appendix;
Figure~\ref{fig:FP_by_cutoff}). Yet, considering the corresponding number of
selected variables per boosting run $q$ (which is inversely related to the
threshold $\thres$), one could see that not only large values of $q$ lead to low
numbers of false positives but also small values (\ref{fig:FP_by_q_and_pinfl}).
This observation is somehow contrary to the optimal choices of $q$ with respect
to the true positive rate. However, an optimal true positive rate is more
important than a low number of false positives as long as the error rate is
controlled.

\section{Case Study: Differential Phenotype Expression for ASD patients versus
  controls}
\label{sec:data-analysis}

We examined autism spectrum disorder (ASD) patients \citep{Manning:ASD:2013} and
compared them to healthy controls. The aim was to detect differentially
expressed amino acid pathways, i.e.\ amino acid pathways that differ between
healthy subjects and ASD patients \citep{Boccuto:ASD:2013}. We used measurements
of absorbance readings from Phenotype Microarrays developed by Biolog (Hayward,
CA). The arrays are designed so as to expose the cells to a single carbon energy
source per well and evaluate the ability of the cells to utilize this energy
source to generate NADH \citep{Bochner:PM:2001}. The array plates were incubated
for 48 h at 37$^\circ$C in 5\% CO2 with 20,000 lymphoblastoid cells per well.
After this first incubation, Biolog Redox Dye Mix MB was added (10 $\mu$L/well)
and the plates were incubated under the same conditions for an additional 24 h.
As the cells metabolize the carbon source, tetrazolium dye in the media is
reduced, producing a purple color according to the amount of NADH generated. At
the end of the 24 h incubation, the plates were analyzed utilizing a microplate
reader with readings at 590 and 750 nm. The first value (A$_{590}$) indicated
the highest absorbance peak of the redox dye and the second value (A$_{750}$)
gave a measure of the background noise. The relative absorbance (A$_{590 -
  750}$) was calculated per well.

Each row of the data set described the measurement of \emph{one well per
  biological replicate}. With $n = 35$ biological replicates ($17$ ASD patients
and $18$ controls) and $p = 4 \cdot 96 = 384$ wells we thus theoretically got
$n\cdot p = 13440$ observations. Due to one missing value the data set finally
contained only $13439$ observations. The data is available as a supplement to
\citet{Boccuto:ASD:2013} and in the \textsf{R} package \textbf{opm}
\citep[G\"{o}ker \citeyear{Goeker:pkg:opm:2014}]{Vaas:2013:opm_bioinf}, which
was also used to store, manage and annotate the data set.

For all available biological replicates we obtained the amino acid annotation
for each measurement in that replicate, i.e.\ we set up an incidence vector per
observation for all available peptides. The incidence vector was one if the
peptide contained that amino acid and zero if it did not. We ended up with $27$
amino acid occurrence annotations in total (including some non-proteinogenic
amino acids). In the next step, we modeled the differences of the measured
values between ASD patients and controls to assess which amino acid pathways were
differentially expressed. Therefore we set up a model of the following form:
\begin{align*}
  \log(y) = \beta_0 + \beta_1 \text{group} & + b_{\text{id}} + \beta_{2,1}
  I_{\text{P1}} +
  \beta_{2,2} I_{\text{P2}} + \ldots + \\
  & + X(\text{group}) \cdot \widetilde{b}_{\text{id}} + X(\text{group}) \cdot
  \beta_{3,1} I_{\text{P1}} + X(\text{group}) \cdot \beta_{3,2} I_{\text{P2}} +
  \ldots,
\end{align*}
where $y$ was the measured PM value, $\beta_0$ was an overall intercept,
$\beta_1$ was the overall group effect (the difference between ASD patients and
controls irrespective of the amino acid that the measurement belonged to).
Additionally, we used an random effect for the replicate ($b_{\text{ID}}$) to
account for subject-specific effects. The amino acid effects $\beta_{2,j}$
represent the differences of the $\log(y)$ values between amino acid, as
$I_{\text{P}j}$ is an indicator function, which was 0 if the well did not belong
to amino acid $j$, and 1 if it did; this means we obtained dummy-coded effect
estimates from the first line of the model formula.

The most interesting part was given by the second line of the model:
$X(\text{group})$ was a group-specific function which was either $-1$ for
controls or $1$ for ASD cases. We used this sum-to-zero constraint in an
interaction with dummy-coded amino acid effects. The coefficients $\beta_{3,j}$
hence represented the deviation of the groups from the global effect of the
$j$th amino acid. If $\beta_{3,j} = 0$, no group-specific effect was present,
i.e.\ the amino acid did not differ between the groups. If $\beta_{3,j} \neq 0$,
the difference between the two groups was twice this effect, i.e.\
$X(\text{ASD}) \cdot \beta_{3,j} - (X(\text{Control}) \cdot \beta_{3,j}) = 1
\cdot \beta_{3,j} - (-1 \cdot \beta_{3,j}) = 2 \beta_{3,j}$. Note that we also
specified a group-specific random effect $\widetilde{b}_{\text{ID}}$.

First, we fitted an offset model containing all main effects, i.e.\ we modeled
differences in the maximum curve height with respect to different amino acids
while neglecting possible differences in amino acid effects between groups. In a
second step, we started from this offset model and additionally allowed for
interactions between the group and the amino acids, while keeping the main
effects in the list of possible base-learners, and checked if any interactions
were present. These represent differential PM expressions between groups.

In total, we ended up with $57$ base-learners (group effect, main amino acid
effects, group-specific effects, and an overall and a group-specific random
effect). All models were fitted using boosting. The selection of differentially
expressed amino acids was done using stability selection. We set the number of
selected variables per boosting model to $q = 10$ and chose an upper bound for
the $\PFER \leq 1$. To judge the magnitude of the multiplicity correction, we
related the used $\PFER$ to the significance level $\alpha$, i.e.\ the standard
$\PCER$: The upper bound for the $\PFER$ equaled $\alpha = 1 / 57 = 0.0175$ in
this setting. With the unimodality assumption, this led to a cutoff $\thres =
0.87$. With the r-concavity assumption, the error bound was $\thres = 0.69$,
while the error bound became $\thres = 1$ without assumptions. Subsequently we
used cross-validation to obtain the optimal stopping iteration for the model.
The code for model fitting and stability selection is given as an electronic
supplement.

\subsection{Results}
\label{sec:results_opm}

\begin{figure}[h]
  \centering
\includegraphics{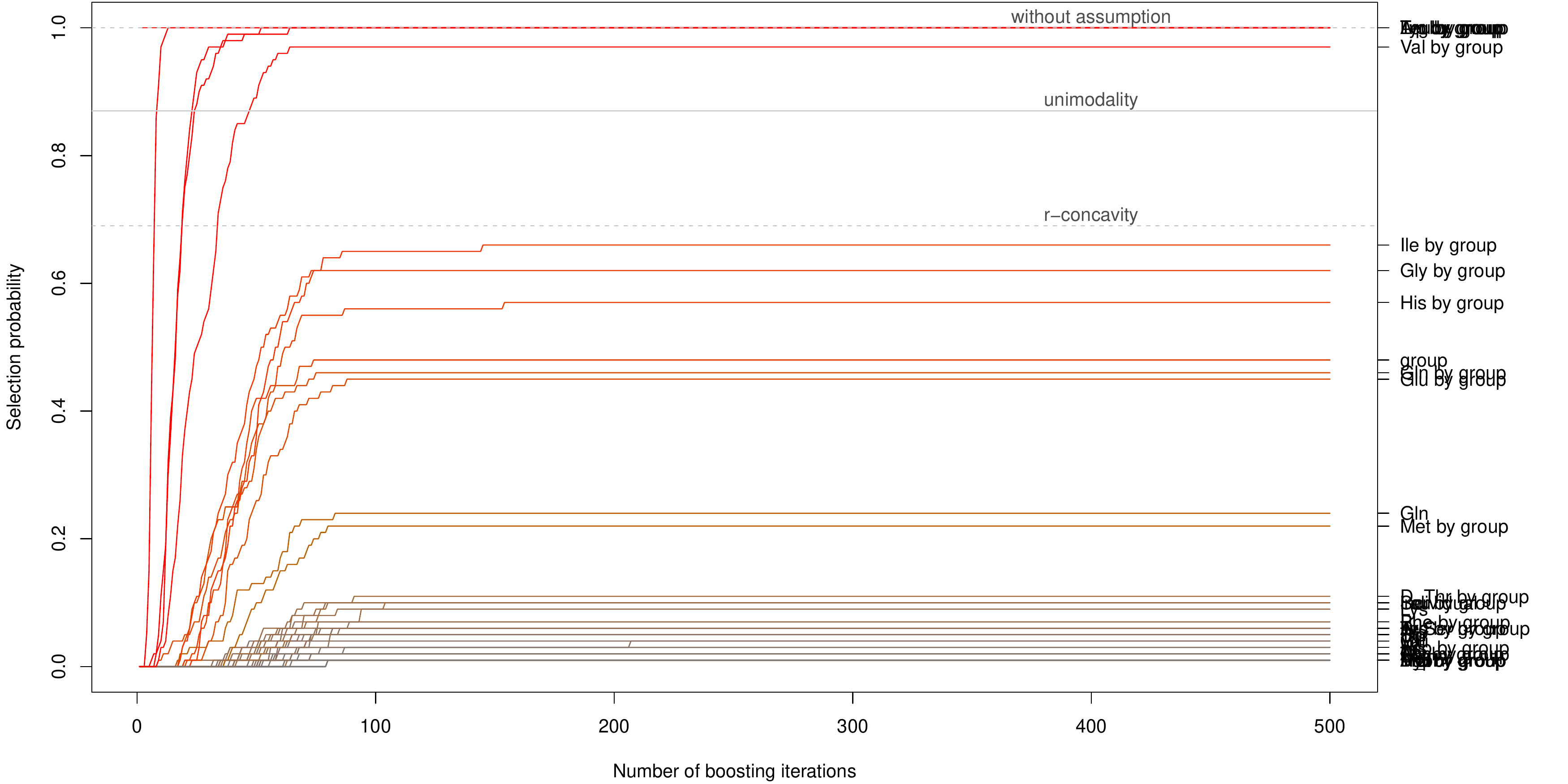}
\caption{Stability selection paths, with the number of boosting iterations
  plotted against the relative selection frequency of the base-learners up to
  that iteration. One can deduce that the number of iterations was sufficiently
  large, as all selection paths cease to increase after approx.\ 150 iterations.
  The solid horizontal gray line is the threshold value with unimodality
  assumption ($\thres = 0.87$), the dashed gray lines represent
  the threshold values with r-concavity assumption ($\thres = 0.69$)
  and without assumption ($\thres = 1$).}
  \label{fig:stabselpaths}
\end{figure}

\begin{figure}[h]
  \centering
\includegraphics{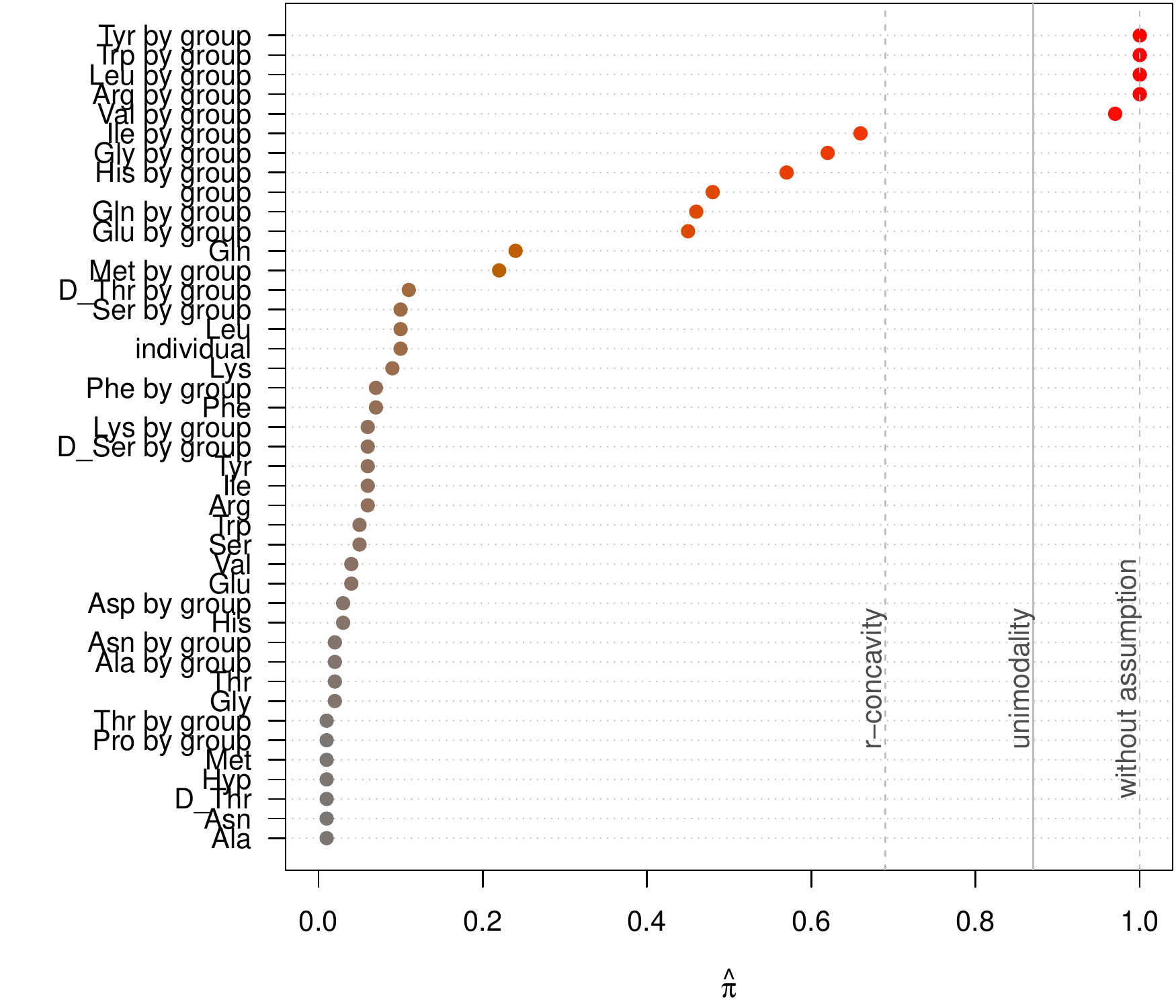}
  \hfill
\includegraphics{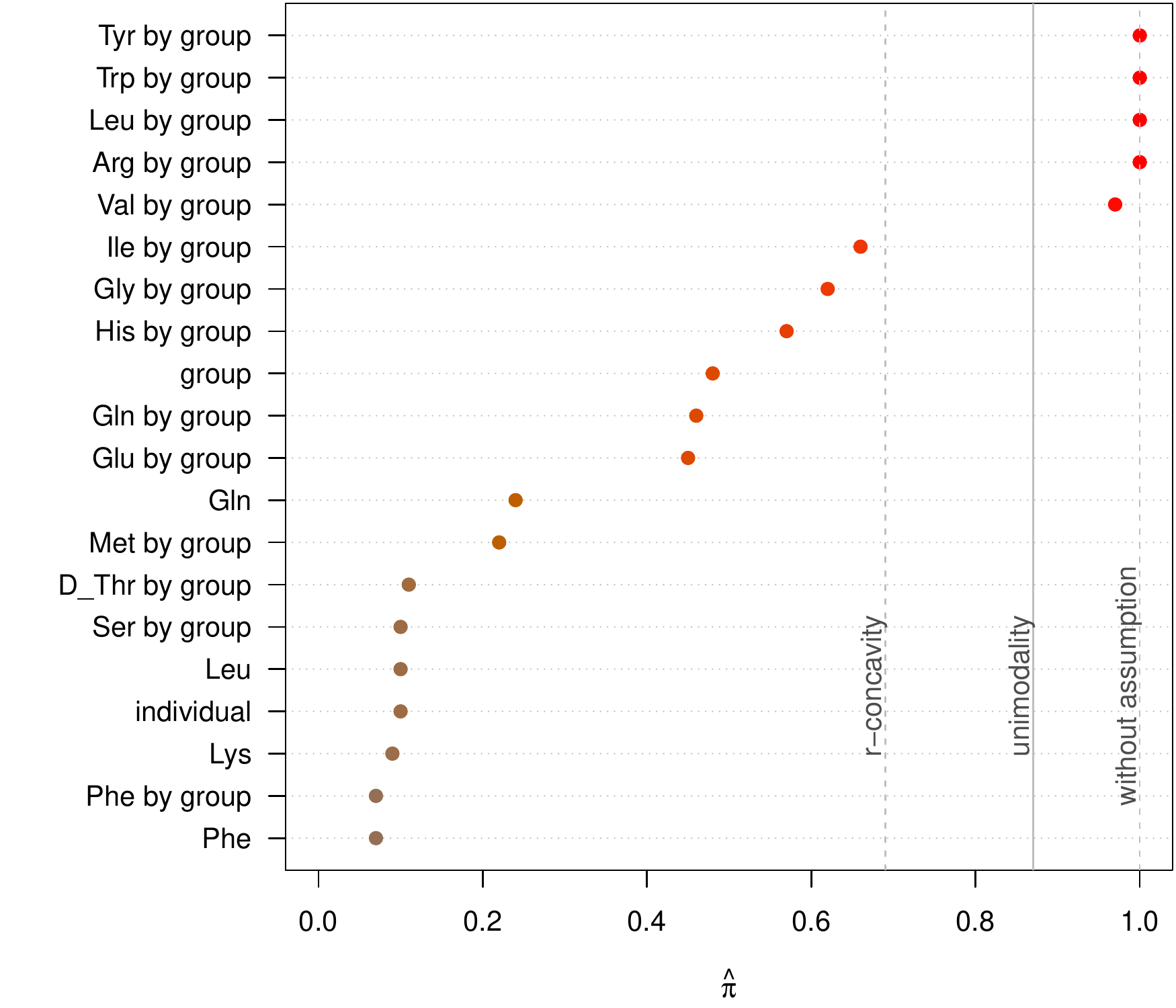}
\caption{The maximum selection frequency $\hat{\pi}$ for all (selected)
  base-learners (left) and for the top 20 base-learners (right) as determined by
  stability selection. The solid vertical gray lines depict the threshold value
  with unimodality assumption ($\thres = 0.87$), the dashed gray
  lines represent the threshold values with r-concavity assumption ($\thres =
  0.69$) and without assumption ($\thres = 1$).}
  \label{fig:stabsel}
\end{figure}

The resulting stability paths can be found in Figure~\ref{fig:stabselpaths}. The
maximum inclusion frequencies for all selected base-learners and for the top
scoring base-learners can be found in Figure~\ref{fig:stabsel}. Tyrosine
(\texttt{Tyr}), tryptophan (\texttt{Trp}), leucine (\texttt{Leu}) and arginine
(\texttt{Arg}) all had a selection frequency of 100 \%. Valine (\texttt{Val})
was selected in 97 \% of the models. Without assumptions, only the
amino acids with 100 \% selection frequency were considered to be stable. Under
the unimodality assumption, valine was additionally termed stable. Together with
the sharp decline in the selection frequency, we would thus focus on these first
five amino acids.

\setkeys{Gin}{width=0.8\textwidth}

\setkeys{Gin}{width=0.45\textwidth}

%%% Local Variables:
%%% mode: latex
%%% TeX-master: "00_main"
%%% End:

The results of our analysis using stability selection confirmed the abnormal
metabolism of the amino acid tryptophan in ASD cells reported by
\cite{Boccuto:ASD:2013}. Additionally, the utilization of other amino acids
seemed to be affected, although on a milder level. When weighted for the size of
the effect, we noticed in ASD patients an overall decreased utilization of
tryptophan ($-0.273$ units on the logarithmic scale), tyrosine ($-0.135$), and
valine ($-0.054$). On the other hand, we registered an increased rate for the
metabolic utilization of arginine ($+0.084$) and leucine ($+0.081$). These
findings suggest an abnormal metabolism of large amino acids (tryptophan,
tyrosine, leucine, and valine), which might be related to impaired transport of
those molecules across the cellular membrane. Separately, a screening by Sanger
sequencing was performed on the coding regions of \emph{SLC3A2}, \emph{SLC7A5},
and \emph{SLC7A8}, the genes coding the subunits of the Large Amino acid
Transporter (LAT) 1 and 2, in 107 ASD patients (including the ones reported in
this paper; Boccuto, unpublished data). Overall, potentially pathogenic
mutations were detected in 17/107 ASD patients ($15.9\%$): eight in
\emph{SLC3A2}, four in \emph{SLC7A5}, and five in \emph{SLC7A8}. We also
evaluated the transcript level for these genes by expression microarray in 10 of
the 17 ASD patients reported in this paper and 10 controls. The results showed
that all the ASD patients had a significantly lower expression of \emph{SLC7A5}
($p$ value $= 0.00627$) and \emph{SLC7A8} ($p$ value $= 0.04067$). Therefore, we
noticed that 27/107 ASD patients ($25.2\%$) had either variants that might
affect the LATs function or reduce the level of transcripts for the
transporters’ subunits. When we correlated the metabolic data collected by the
Phenotype Microarrays with those findings, we noticed that all of these patients
showed reduced utilization of tryptophan. Additionally, eight out of the twelve
patients who were screened with the whole metabolic panel showed significantly
reduced tyrosine utilization in at least 25 of the 27 wells containing this
amino acid, seven had a reduced utilization of valine in at least 29/34 wells,
and five had a reduced metabolism of leucine in at least 27/31 wells. These data
are concordant with the present findings as they suggest an overall problem with
the metabolism of large amino acids, which might have important consequences in
neurodevelopment and synapsis homeostasis, especially if one considers that such
amino acids are precursors of important compounds, such as serotonin, melatonin,
quinolinic acid, and kynurenic acid (tryptophan), or dopamine (tyrosine).

\section{Discussion}
\label{sec:discussion}

Stability selection proofs to work well in high-dimensional settings with (much)
more predictors than observations. It adds an error control to the selection
process of boosting or other high-dimensional variable selection approaches.
Assumptions on the distribution of the simultaneous selection probabilities
increase the number of true positive variables, while keeping the error control
in most settings. As shown in our case study, complex log-linear interaction
models can be used as learners in conjunction with stability selection.
Additionally, more complex models such as generalized additive models
\citep[GAMs; ][]{Hast:Tibs:gene:1986,hastie:tib:gams:1990} or structured
additive regression (STAR) models \cite{Fahrmeir2004,Hofner:constrained:2014}
can also benefit from the combination with stability selection if model or
variable selection (with a control for the number of false positives) is of
major interest.

However, one should keep in mind that stability selection controls the
per-family error rate, which is very conservative. Specifying the error rate
such that $\alpha \leq \PFERmax \leq m \alpha$, with significance level $\alpha$
and $m$ hypothesis tests, might provide a good idea for a sensible error control
in high-dimensional settings with $\FWER$-control ($\PFERmax =\alpha$) and no
multiplicity adjustment ($\PFERmax = m \alpha$) as the extreme cases.

Furthermore, prediction models might not always benefit from stability
selection. If the error control is tight, i.e.\ $\PFERmax$ is small, the true
positive rate is usually smaller than in a cross-validated prediction model
without stability selection and the prediction accuracy suffers \citep[see
also][]{Hothorn:stabsel:2010}. Prediction and variable selection are two
different goals.

\subsection*{Implementation}

The component-wise, model-based boosting approach is implemented in the
\textsf{R} add-on package \textbf{mboost}
\citep{buehl:hoth:2007,Hothorn+Buehlmann+Kneib+Schmid+Hofner:mboost:2010,pkg:mboost:CRAN:2.4}.
A comprehensive tutorial for \textbf{mboost} is given in
\citet{Hofner:mboost:2014}. The \textsf{R} package \textbf{opm} \citep[G\"{o}ker
\citeyear{Goeker:pkg:opm:2014}]{Vaas:2013:opm_bioinf} is used to store, manage
and annotate the data set. Tutorials are given as vignettes.

Stability selection is implemented in the add-on package \textbf{stabs}
\citep{pkg:stabs:CRAN:0.1} for the statistical program environment \textsf{R}
\citep{r-core:2014}. One can directly use stability selection on a fitted
boosting model using the function \texttt{stabsel}. One only needs to
additionally specify two of the parameters \texttt{PFER}, \texttt{cutoff} and
\texttt{q}. The missing parameter is then computed such that the specified type
of error bound holds (without additional assumptions (E\ref{item:errorbound1}),
under unimodality (E\ref{item:errorbound2}) or under r-concavity
(E\ref{item:errorbound3})). Alternative \texttt{stabsel} methods exist for
various other fitting approaches (e.g. Lasso) using a matrix or a formula
interface. By specifying a function that returns the index (and names) of
selected variables one can easily extend this framework. In general, the
function \texttt{stabsel\_parameters} can be used to compute the missing
parameter without running stability selection itself to check if the value of
the parameter computed from the other two parameters is sensible in the data
situation at hand. 

\subsection*{Acknowledgments}

We thank Rajen D. Shah, N. Meinshausen and P. B\"{u}hlmann for helpful comments
and discussion, Chin-Fu Chen and Charles E. Schwartz from the Greenwood Genetic
Center for their help with the analysis and with the interpretation of the
results, as well as Michael Drey who conducted an early version of the presented
simulation study. 

\bibliographystyle{jss_modified}
\bibliography{bibliography_full}

\clearpage

\appendix

\section{Additional Figures}
\label{sec:additional-figures}

\setkeys{Gin}{width = 0.79\textwidth}

\begin{figure}[h]
  \centering
\includegraphics{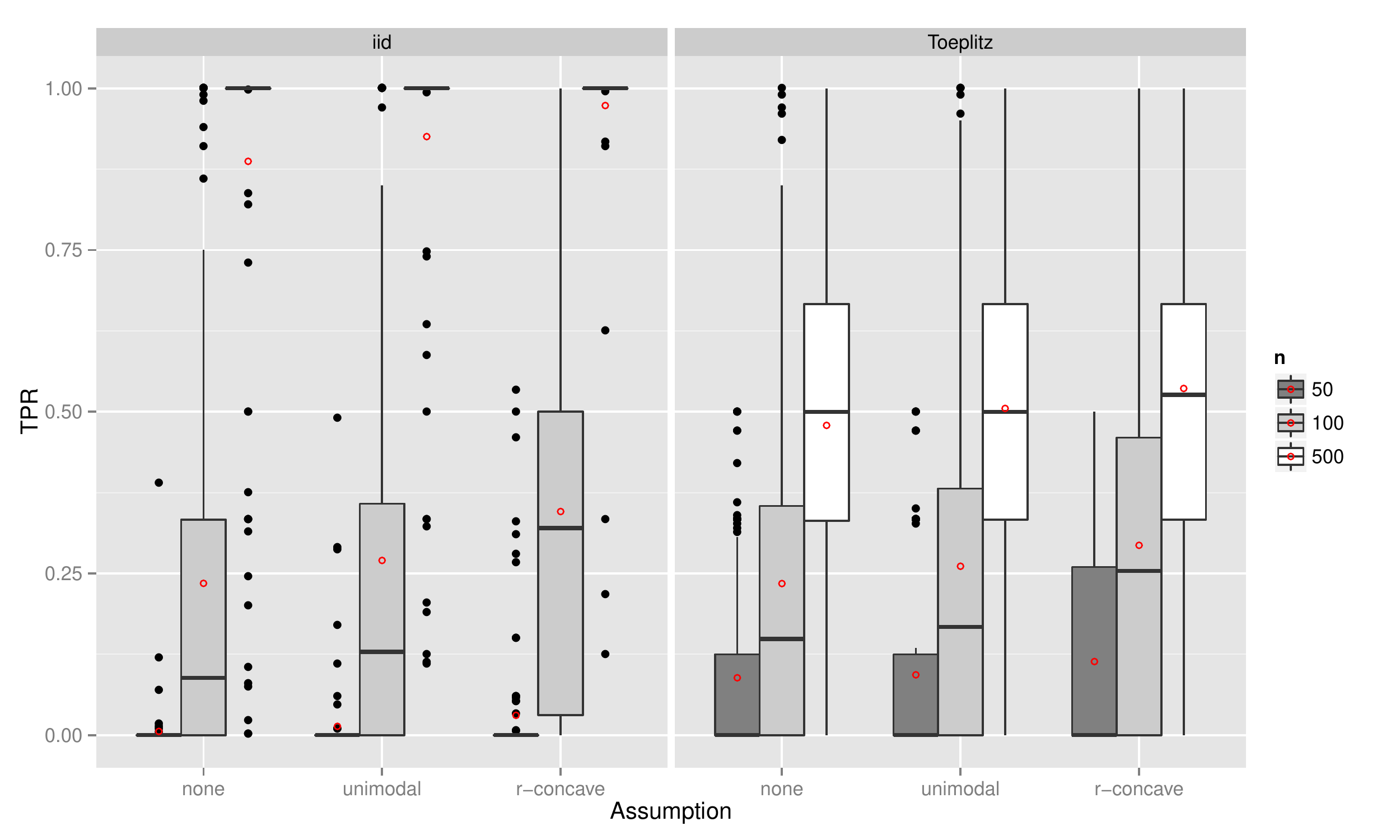}
\caption{Boxplots for the true positives rates (TPR) for all simulation settings
  with separate boxplots for different numbers of observations ($n$), the
  correlation settings (independent predictor variables or Toeplitz design), and
  the assumptions used to compute the error bound. Each observation in the
  boxplot is the average of the 50 simulation replicates. The open red circles
  represent the average true positive rates.}
  \label{fig:TPR_by_n}
\end{figure}

%%% Local Variables:
%%% mode: latex
%%% TeX-master: "00_main"
%%% End:

\hfill

\begin{figure}[h]
  \centering
\includegraphics{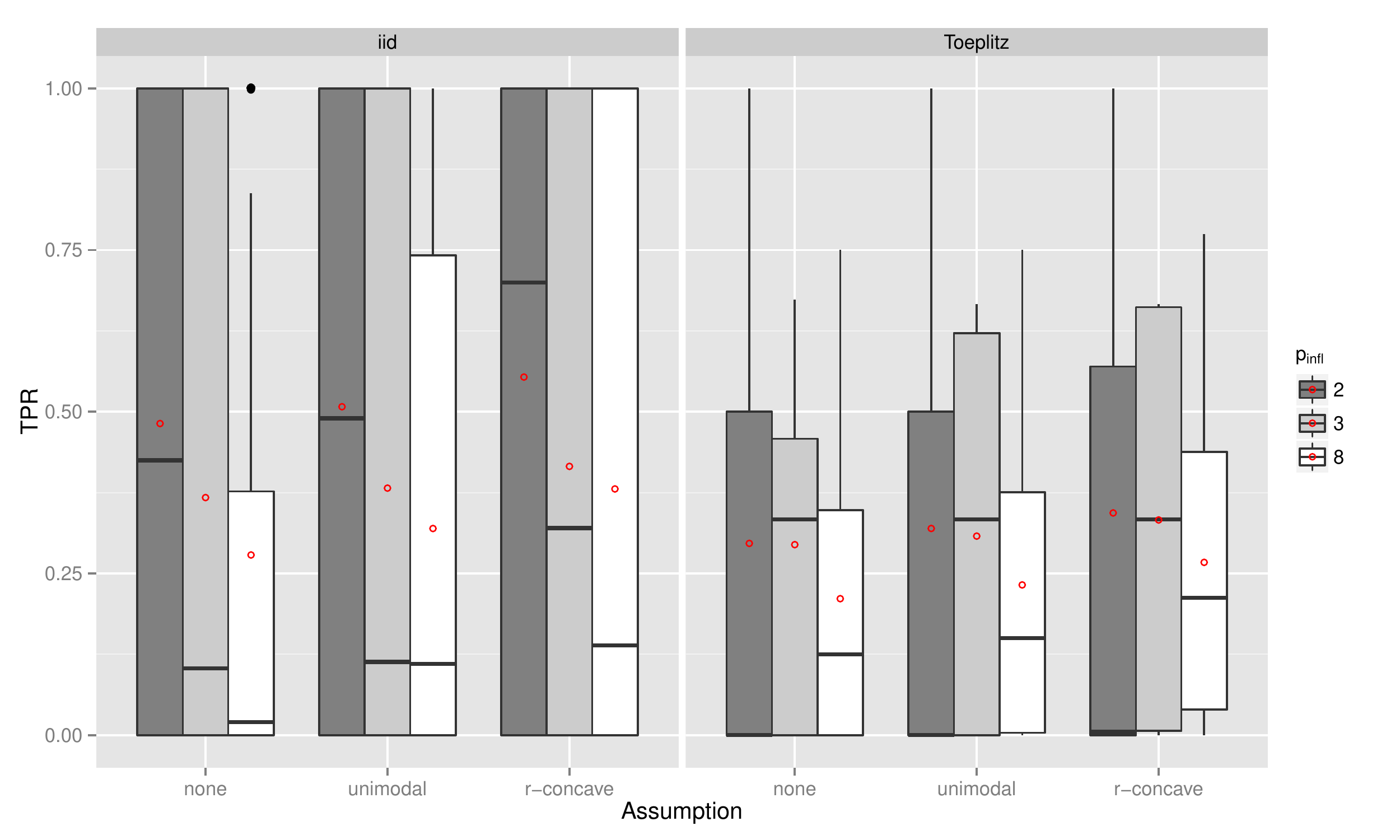}
\caption{Boxplots for the true positives rates (TPR) for all simulation settings
  with separate boxplots for different numbers of influential variables
  ($p_{\text{infl}}$), the correlation settings (independent predictor variables
  or Toeplitz design), and the assumptions used to compute the error bound. Each
  observation in the boxplot is the average of the 50 simulation replicates. The
  open red circles represent the average true positive rates.}
  \label{fig:TPR_by_pinfl}
\end{figure}

%%% Local Variables:
%%% mode: latex
%%% TeX-master: "00_main"
%%% End:

\begin{figure}[h]
  \centering
\includegraphics{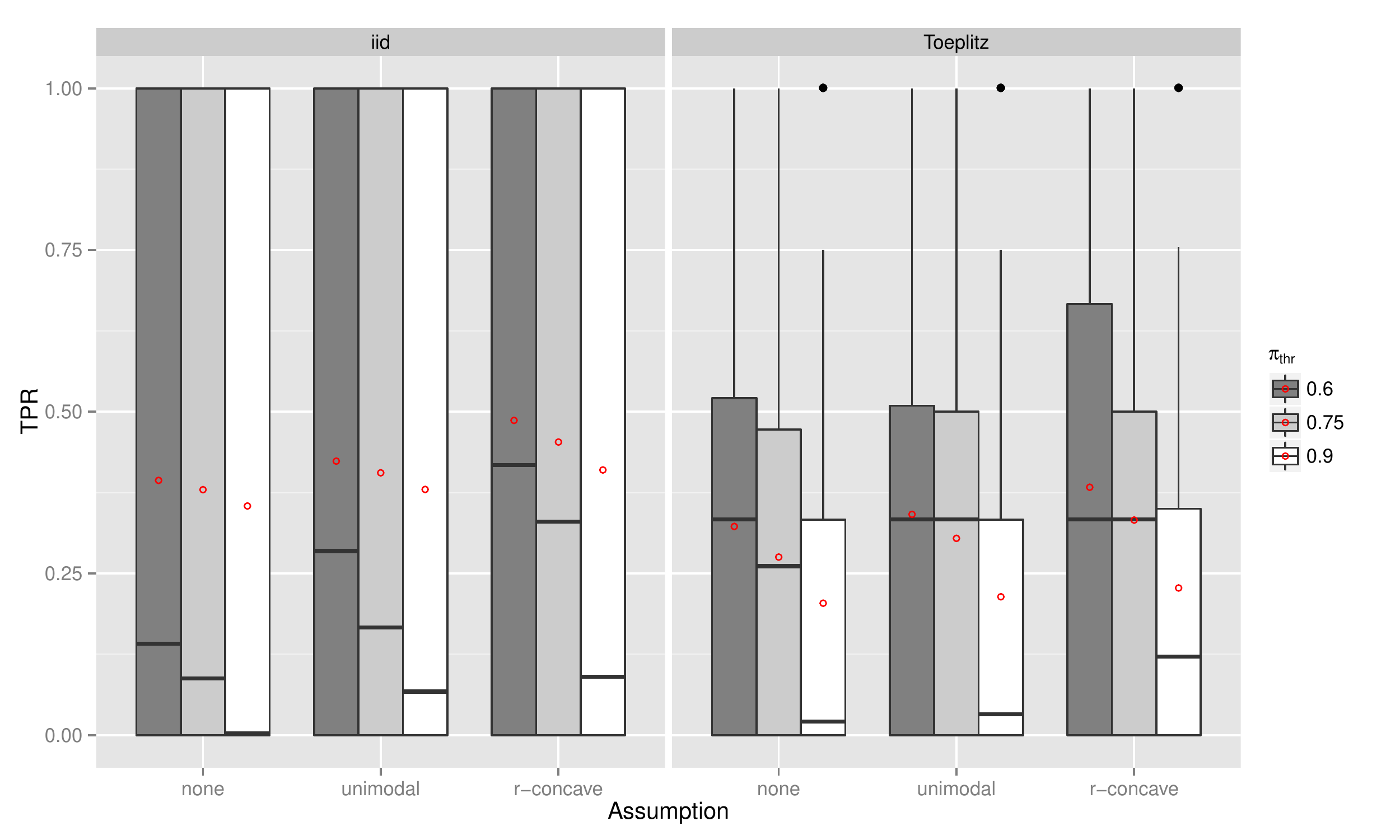}
\caption{Boxplots for the true positives rates (TPR) for all simulation settings
  with separate boxplots for different cutoff values ($\thres$), the correlation
  settings (independent predictor variables or Toeplitz design), and the
  assumptions used to compute the error bound. Each observation in the boxplot
  is the average of the 50 simulation replicates. The open red circles represent
  the average true positive rates.}
  \label{fig:TPR_by_cutoff}
\end{figure}

%%% Local Variables:
%%% mode: latex
%%% TeX-master: "00_main"
%%% End:

\hfill

\begin{figure}[h]
  \centering
\includegraphics{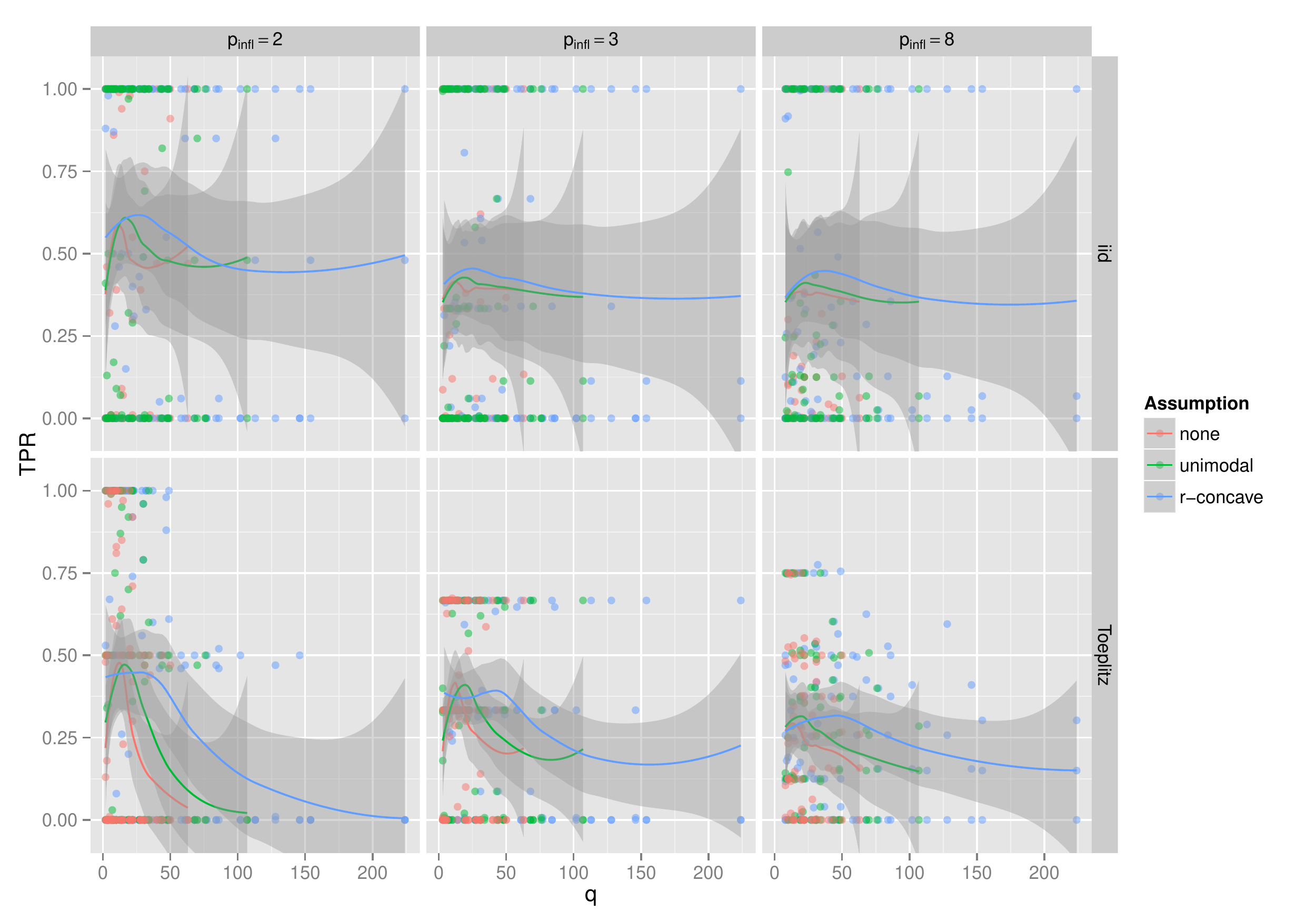}
\caption{Scatter plot showing the true positives rates (TPR) for all simulation
  settings where $q$ was larger than the number of influential variables
  ($p_{\text{infl}}$); Plots are shown separately for the number of influential
  variables ($p_{\text{infl}}$), the correlation settings (independent predictor
  variables or Toeplitz design) and the assumptions used to compute the error
  bound. Each observation in the plot is the average of the 50 simulation
  replicates. The lines depict a scatter plot smoother for each group together
  with shaded confidence regions. \label{fig:TPR_by_q_and_pinfl}}
\end{figure}

%%% Local Variables:
%%% mode: latex
%%% TeX-master: "00_main"
%%% End:

\begin{figure}[h]
  \centering
\includegraphics{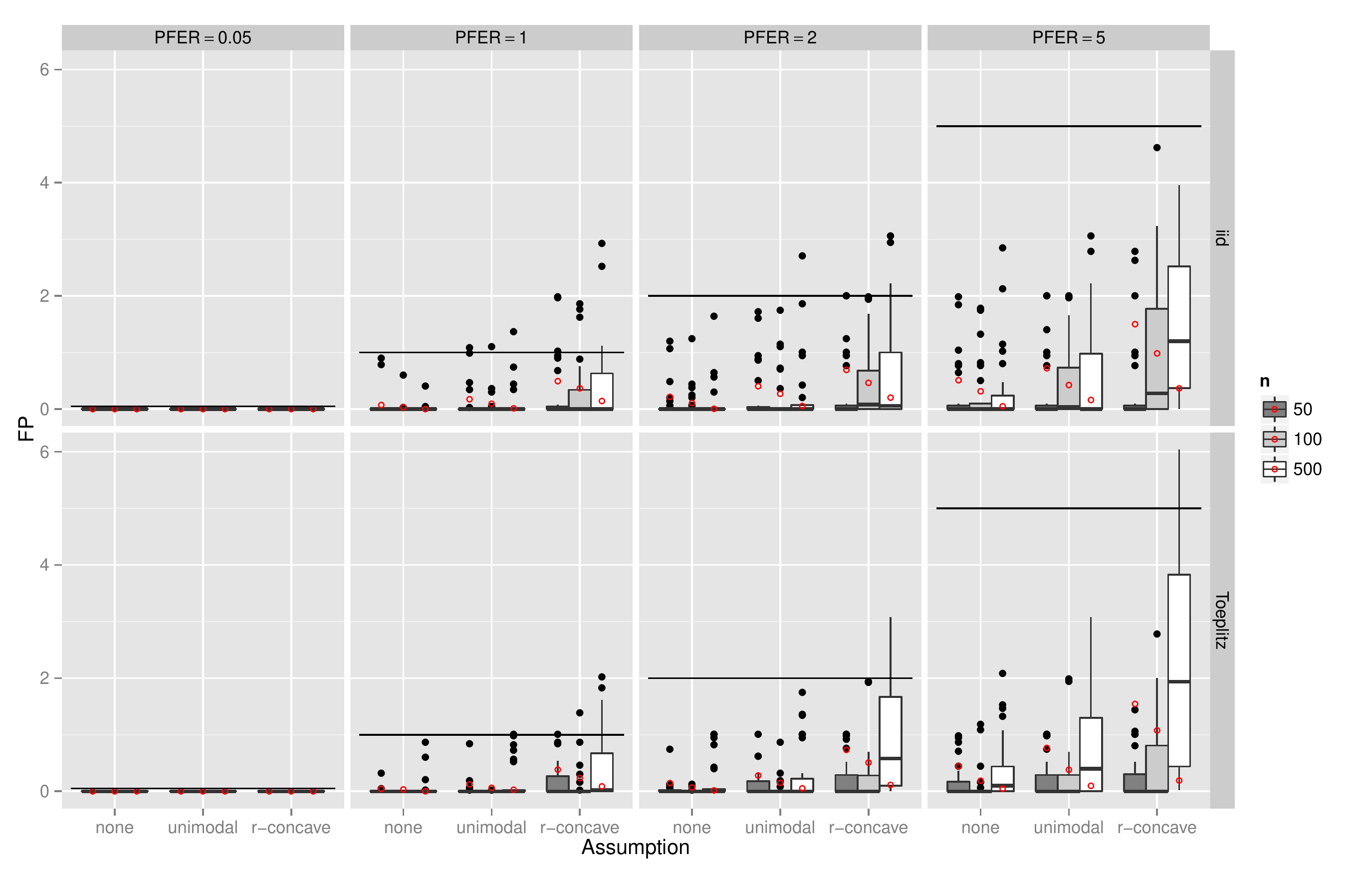}
\caption{Boxplots for the number of false positives (FP) for all simulation
  settings with separate boxplots for different numbers of observations ($n$),
  the correlation settings (independent predictor variables or Toeplitz design),
  the $\PFER$, and the assumptions used to compute the error bound. Each
  observation in the boxplot is the average of the 50 simulation replicates. The
  open red circles represent the average number of false positives.}
  \label{fig:FP_by_n}
\end{figure}

%%% Local Variables:
%%% mode: latex
%%% TeX-master: "00_main"
%%% End:

\hfill

\begin{figure}[h]
  \centering
\includegraphics{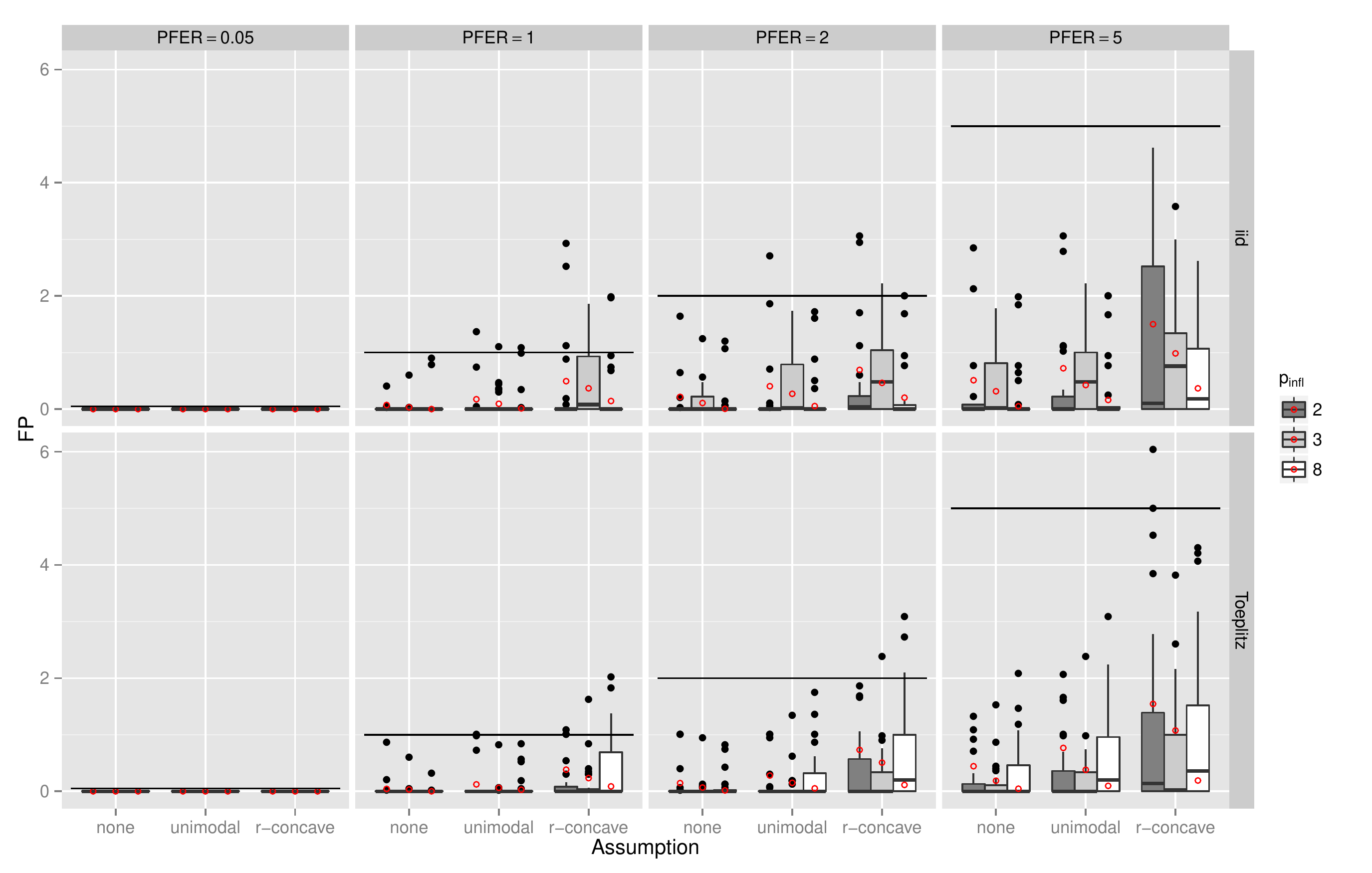}
\caption{Boxplots for the number of false positives (FP) for all simulation
  settings with separate boxplots for different numbers of influential variables
  ($p_{\text{infl}}$), the correlation settings (independent predictor variables
  or Toeplitz design), the $\PFER$, and the assumptions used to compute the
  error bound. Each observation in the boxplot is the average of the 50
  simulation replicates. The open red circles represent the average number of
  false positives.}
  \label{fig:FP_by_pinfl}
\end{figure}

%%% Local Variables:
%%% mode: latex
%%% TeX-master: "00_main"
%%% End:

\begin{figure}[h]
  \centering
\includegraphics{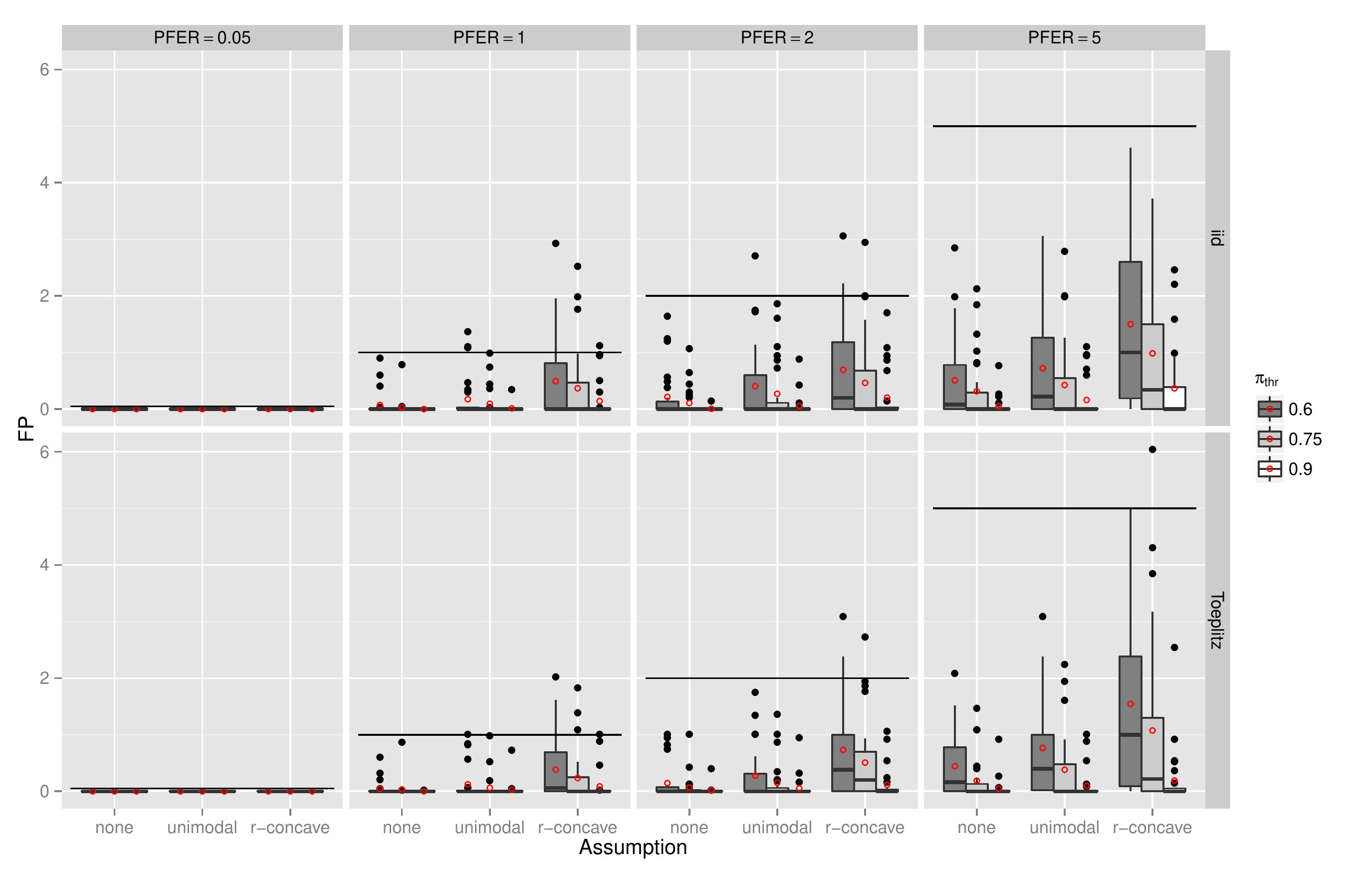}
\caption{Boxplots for the number of false positives (FP) for all simulation
  settings with separate boxplots for different cutoff values ($\thres$), the
  correlation settings (independent predictor variables or Toeplitz design), the
  $\PFER$, and the assumptions used to compute the error bound. Each observation
  in the boxplot is the average of the 50 simulation replicates. The open red
  circles represent the average number of false positives.}
  \label{fig:FP_by_cutoff}
\end{figure}

%%% Local Variables:
%%% mode: latex
%%% TeX-master: "00_main"
%%% End:

\hfill

\begin{figure}[h]
  \centering
\includegraphics{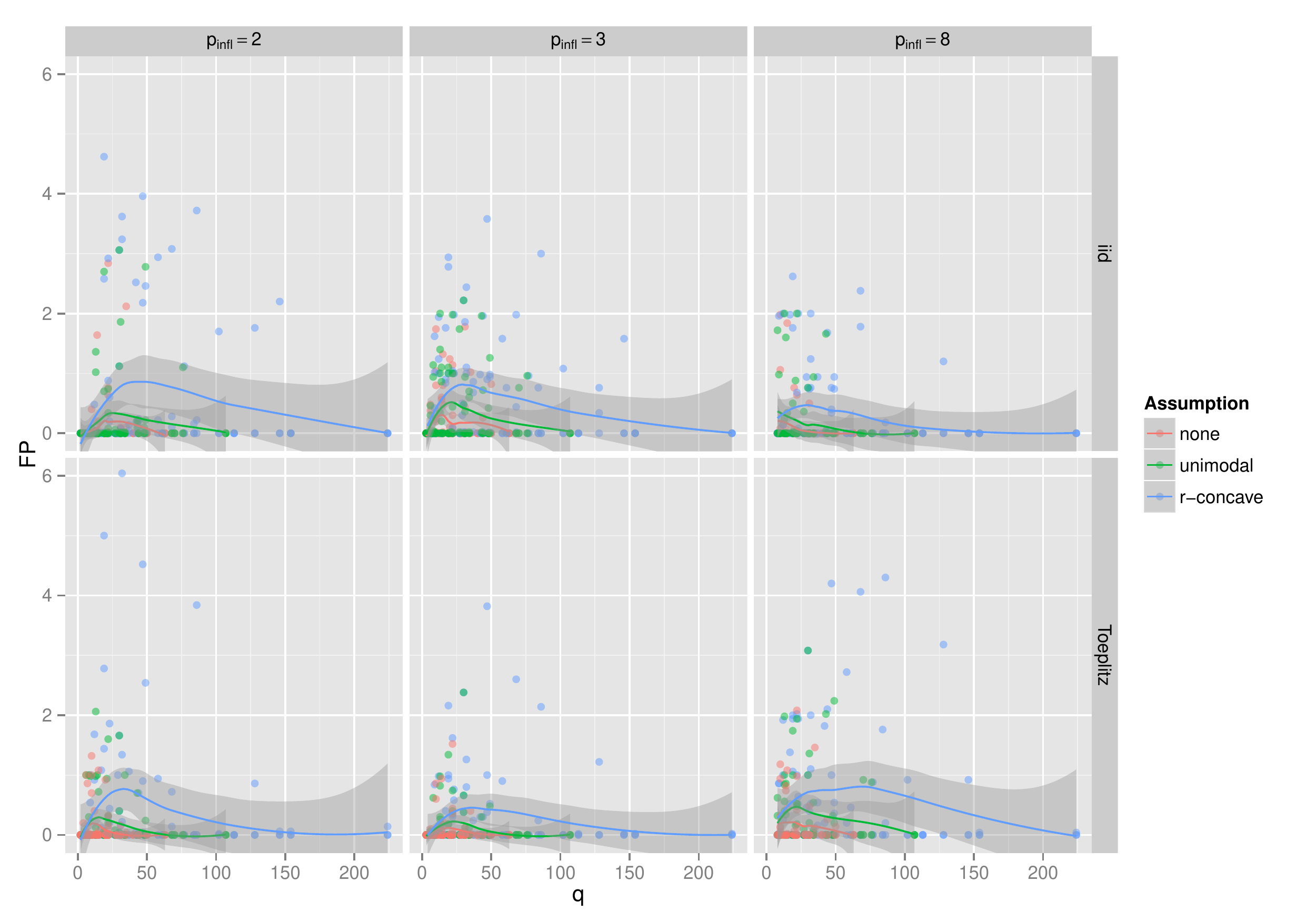}
\caption{Scatter plot showing the number of false positives (FP) for all
  simulation settings where $q$ was larger than the number of influential
  variables ($p_{\text{infl}}$); Plots are shown separately for the number of
  influential variables ($p_{\text{infl}}$), the correlation settings
  (independent predictor variables or Toeplitz design) and the assumptions used
  to compute the error bound. Each observation in the plot is the average of the
  50 simulation replicates. The lines depict a scatter plot smoother for each
  group together with shaded confidence regions. \label{fig:FP_by_q_and_pinfl}}
\end{figure}

%%% Local Variables:
%%% mode: latex
%%% TeX-master: "00_main"
%%% End:

\end{document}